%% file: main.tex
\documentclass{article}

\PassOptionsToPackage{numbers, compress}{natbib}
    \usepackage[preprint]{neurips_2024}

\usepackage[utf8]{inputenc} % allow utf-8 input
\usepackage[T1]{fontenc}    % use 8-bit T1 fonts
\usepackage{hyperref}       % hyperlinks
\usepackage{url}            % simple URL typesetting
\usepackage{booktabs}       % professional-quality tables
\usepackage{amsfonts}       % blackboard math symbols
\usepackage{nicefrac}       % compact symbols for 1/2, etc.
\usepackage{microtype}      % microtypography
\usepackage{xcolor}         % colors
\usepackage{natbib}
\usepackage{amsmath}
\usepackage{graphicx}
\usepackage{subcaption}
\usepackage{amsthm}

\newtheorem{theorem}{Theorem}

\newcommand{\ie}{\textit{i}.\textit{e}.}
\newcommand{\eg}{\textit{e}.\textit{g}.}

\input{math_commands}

\newcommand{\Span}{\mathit{Span}}

\title{PaRa: Personalizing Text-to-Image Diffusion via Parameter Rank Reduction}

\author{Shangyu Chen,~~Zizheng Pan,~~Jianfei Cai,~~Dinh Phung\\
Monash University, Melbourne, Australia\\
\texttt{\{shangyu.chen,zizheng.pan,jianfei.cai,dinh.phung\}@monash.edu}}

\begin{document}

\maketitle

\begin{abstract}
  Personalizing a large-scale pretrained Text-to-Image (T2I) diffusion model is challenging as it typically struggles to make an appropriate trade-off between its training data distribution and the target distribution, \ie, learning a novel concept with only a few target images to achieve personalization (aligning with the personalized target) while preserving text editability (aligning with diverse text prompts).  
  % The diffusion models have made significant progress in generating images from text, facilitating the creation of high-quality visuals from textual prompts or other input modalities. However, existing methods for customizing diffusion models always face the problem of the balance between the diversity of generated images and the faithfulness to the customization objectives. 
  % As a result, while they perform well in customizing single-subject generation, they may be limited in more advanced tasks such as mixing between different customized concepts or image editing. 
  In this paper, we propose \textbf{PaRa}, an effective and efficient \textbf{Pa}rameter \textbf{Ra}nk Reduction approach for T2I model personalization by explicitly controlling the rank of the diffusion model parameters to restrict its initial diverse generation space into a small and well-balanced target space. Our design is motivated by the fact that taming a T2I model toward a novel concept such as a specific art style implies a small generation space. To this end, by reducing the rank of model parameters during finetuning, we can effectively constrain the space of the denoising sampling trajectories towards the target.
  % where we shift the focus from the values of weights in the diffusion models to the rank of weights, thus customizing the output space of the diffusion models to subspaces, this allows for a more precise learning of customized concepts. 
  % Our proposed PaRa method has a smaller model size compared to parameter-efficient methods like LoRA, making it more practical for real-world applications, and can be mathematically proven.
  With comprehensive experiments, we show that PaRa achieves great advantages over existing finetuning approaches on single/multi-subject generation as well as single-image editing. Notably, compared to the prevailing fine-tuning technique LoRA, PaRa achieves better parameter efficiency ($2\times$ fewer learnable parameters) and much better target image alignment. 
  % \zizheng{double-check this claim, considering Figure~\ref{fig:text_image_alignment}. } 
\end{abstract}

\input{sections/introduction}

\input{sections/related}

\input{sections/method}

\input{sections/experiments}

\input{sections/conclusion}

\newpage

% \section*{References}

% References follow the acknowledgments in the camera-ready paper. Use unnumbered first-level heading for
% the references. Any choice of citation style is acceptable as long as you are
% consistent. It is permissible to reduce the font size to \verb+small+ (9 point)
% when listing the references.
% Note that the Reference section does not count towards the page limit.
% \medskip

\bibliographystyle{plain}
\bibliography{main} % This should match the name of your .bbl file

% {\small  % reducing font size as recommended
% \bibliography{reference} %.bib?
% } 

% {
% \small

% [1] Alexander, J.A.\ \& Mozer, M.C.\ (1995) Template-based algorithms for
% connectionist rule extraction. In G.\ Tesauro, D.S.\ Touretzky and T.K.\ Leen
% (eds.), {\it Advances in Neural Information Processing Systems 7},
% pp.\ 609--616. Cambridge, MA: MIT Press.

% [2] Bower, J.M.\ \& Beeman, D.\ (1995) {\it The Book of GENESIS: Exploring
%   Realistic Neural Models with the GEneral NEural SImulation System.}  New York:
% TELOS/Springer--Verlag.

% [3] Hasselmo, M.E., Schnell, E.\ \& Barkai, E.\ (1995) Dynamics of learning and
% recall at excitatory recurrent synapses and cholinergic modulation in rat
% hippocampal region CA3. {\it Journal of Neuroscience} {\bf 15}(7):5249-5262.
% }

%%%%%%%%%%%%%%%%%%%%%%%%%%%%%%%%%%%%%%%%%%%%%%%%%%%%%%%%%%%%
\newpage
\appendix

\input{sections/appendix}

% Optionally include supplemental material (complete proofs, additional experiments and plots) in appendix.
% All such materials \textbf{SHOULD be included in the main submission.}

%%%%%%%%%%%%%%%%%%%%%%%%%%%%%%%%%%%%%%%%%%%%%%%%%%%%%%%%%%%%

\end{document}

%% file: math_commands.tex
\def\0{{\bf 0}}
\def\1{{\bf 1}}

%% file: sections/introduction.tex
\section{Introduction}
\label{sec:PaRa_intro}

Recent text-to-image (T2I) diffusion models~\citep{sd,sdxl,dalle3,midjourney} 
have achieved unprecedented success. However, despite being trained on large-scale datasets, most T2I models struggled to generate novel concepts as they were limited within their training data distribution. For example, %it is quite obvious that 
pretrained Stable Diffusion (SD) models~\cite{sd} cannot generate unseen objects like a novel anime character. Thus, it has drawn increasing attention in the community to teach off-the-shelf T2I models with a few target images to learn a novel concept (\eg, a specific personal pet) for aligning with user preferences, which is known as T2I model personalization.

Much effort has been made in this direction. Some works~\cite{jia2023taming,wei2023elite} seek for a general encoder to learn a novel concept without test-time finetuning. However, training such an encoder usually requires building a large collection of text-image pairs and expensive computing resources, \eg, 1.4 million pairs for the category of person in InstantBooth~\cite{shi2023instantbooth} and 128 TPUv4 chips in SuTI~\cite{chen2024subject}. Another prevalent line of work directly finetunes T2I models based on the target images, which can be roughly categorized into using text embedding~\cite{textinversion}, cross-attention layers~\cite{kumari2023multi}, full model finetuning~\cite{dreambooth}, low-rank update~\cite{hu2021lora,lora_git} or adjusting singular values of model parameters~\cite{han2023svdiff}. However, all these fine-tuning methods directly change the initial full generation space, which naturally results in a trade-off between generation diversity and the alignment with the target concept. Consequently, they usually suffer from either poor alignment on the new concept~\cite{textinversion}, or overfitting the few target text-image pairs~\cite{dreambooth}.

In this paper, we introduce PaRa, a new parameter-efficient framework for T2I model personalization via \textbf{Pa}rameter \textbf{Ra}nk Reduction. Our motivation comes from two folds: First, diffusion models are trained to capture their training data distribution, indicating a diverse image generation space due to large-scale pretraining. However, taming a T2I model for a novel concept may instead suggest a small restricted generation space. For example, finetuning SDs into a specific anime style like Ghibli Studio\footnote{https://huggingface.co/nitrosocke/Ghibli-Diffusion} does not requires strong photorealistic knowledge learned from LAION-5B dataset~\cite{laion5b}. Second, in a broad literature, the rank of matrix has been shown to be effective to control the degree of certain restrictions such as in image compression~\cite{andrews1976singular}, network pruning~\cite{IdelbayevC20} and parameter-efficient finetuning~\cite{hu2021lora}.

Based on the discussions above, our key idea is to explicitly control the rank of a layer output in a diffusion model during the denoising sampling, thus effectively steering the image generation into a well-aligned and restricted space for learning a new concept. 
In practice, this is achieved by introducing a low-rank learnable parameter $B \in \mathbb{R}^{d \times r}$ for a pretrained linear projection $W_0 \in \mathbb{R}^{d \times k}$, where $r$ is the reduced rank and $d$ is the number of hidden dimensions of this layer. Next, with a simple QR decomposition of $B  = QR$, we can obtain a set of orthonormal bases $Q$ which can be adopted to reduce the rank of the initial layer outputs by $Wx - QQ^T Wx$, where $x$ denotes the layer input. We theoretically prove this result in Appendix \ref{sec:appendixA}.
% Section~\ref{}. \zizheng{@shangyu, please add this after finishing appendix} 
Note that this is fundamentally different from LoRA ~\citep{lora_git,hu2021lora},  which adds low-rank updates to a layer output to achieve parameter-efficienct finetuning.

\begin{figure}[]
\centering            \includegraphics[width=1.0\linewidth]{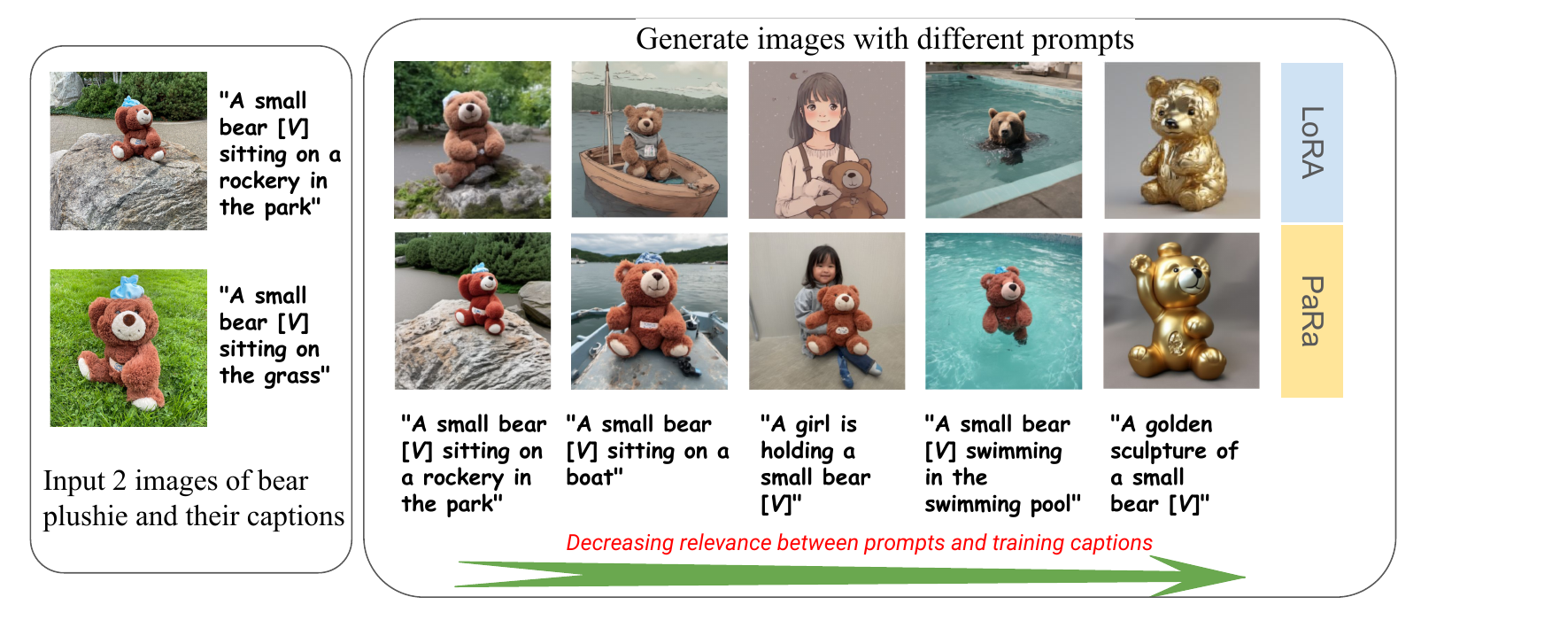}
\caption{Comparison between LoRA ~\citep{lora_git} and our propsoed PaRa on T2I personalization for learning a new concept of bear, \ie, ``\textit{[V]}''.
% Personalized images generated by LoRA and our model given different prompts. Based on a small number of images and caption combinations as few-shot training inputs. 
 For a fair comparison, we set the rank as 4 and adopt the same latent noise for both methods.
% Both LoRA and PaRa adopts rank=4 and were generated under the same random seed. 
LoRA scale is set to 1.0.
}
\label{fig:intro}
\end{figure}

With extensive experiments, we demonstrate several advantages of PaRa. First, compared to the typical LoRA fine-tuning \citep{lora_git}, PaRa can generate images far beyond its initial training prompts while maintaining well the consistency of the specific target concept, as shown in Figure~\ref{fig:intro}. Furthermore, PaRa only requires half of the storage cost of LoRA since it only needs one matrix $B$ for each layer.  Second, benefiting from the explicit rank control, 
% \zizheng{@shangyu, please help to double-check why we are better in this setting. I wrote this but not 100\% sure.}
we further propose an approach to combine multiple individually fine-tuned PaRa weights, %where it demonstrates well generation quality on 
which enables the blending
of multiple personalized concepts for multi-subject T2I generation. %can be used for generating images with multiple target subjects, 
% without expensive training with dedicated data augmentation strategies.
without the need for additional training with augmented multi-subject images~\cite{han2023svdiff}. Third, PaRa is fully compatible with LoRA weights~\cite{hu2021lora}, supporting the combination of PaRa with a wide range of existing LoRA finetuned SD models in the public model zoo. Finally, thanks to the restricted small generation space, single image editing under PaRa is quite stable as the initial noise has a negligible effect on the final image generation. For example, as demonstrated in the experimental section~\ref{Single_Image_Editing}, under the same prompts, PaRa achieves a significantly higher average SSIM, compared to the baseline. %, indicating more stable image generation. 
% \zizheng{xxx, concrete experiment results to prove our advantage.}

%To sum up, we demonstrate our contributions as following.
The major contributions of this work can be summarized as follows.
\begin{itemize}
    \item We propose Parameter Rank Reduction (PaRa), a new framework for personalizing T2I diffusion models by explicitly reducing the rank of the diffusion model parameters.
    \item We propose a simple yet effective approach to combine multiple individually fine-tuned PaRa models, enabling the blending of multiple personalized concepts. Our model also facilitates single-image editing without the need to do the noise inversion~\citep{song2020denoising,mokady2023null} %trace the noise 
    in the diffusion process.
    \item We conduct comprehensive experiments to show that PaRa achieves state-of-the-art performance in personalized single/multi-subject generation.
\end{itemize}

%% file: sections/related.tex
\section{Related work}
\label{sec:PaRa_related}
\subsection{Text-to-Image Diffusion Models}
%The evolution of 
Generative models exemplified by Diffusion Models\citep{chang2023muse,gu2022vector,ho2020denoising,nichol2021improved,song2020denoising,sohl2015deep,song2020score} have achieved significant advancements in the past few years. 
% \jf{You could at least add a few lines here to summarize the existing T2I diffusion models. Pls use ChatGPT to polish your writing first.} %Presently, transformer-based architectures are integral to the standard configuration of Diffusion Models. The experiments in this paper utilize 
\citet{ho2020denoising} proposed the Denoising Diffusion Probabilistic Models (DDPM), which enable diffusion models to achieve excellent performance in image generation tasks. \citet{song2020denoising} further improved this approach by combining score function with diffusion probabilistic models, significantly enhancing the efficiency of image generation. 
DALL-E 2 \citep{ramesh2022hierarchical}, Imagen \citep{saharia2022imagen}, and Stable Diffusion \citep{sd} iteratively denoise data within a latent space, trained on large image-text datasets, significantly enhancing the practicality and effectiveness of diffusion models.

Among the existing T2I diffusion models, Stable Diffusion (SD) \citep{sd} 
% \jf{Give a reference} 
is the most widely used one, which is a variant of Latent Diffusion Models (LDMs) \citep{sd}. LDMs encode input images $x$ into a latent code $z$ with encoder $\mathcal{E}$, and then execute the denoising process on $z$. The training objective is to minimize
\begin{align} 
\mathbb{E}_{\mathcal{E}(x), y, \epsilon \sim \mathcal{N}(0,1), t}\left[\left\|\epsilon-\epsilon_\theta\left(z_t, t, \tau(y)\right)\right\|_2^2\right]
\end{align} 
where $y$ is the input condition such as text or semantic maps, $\tau$ is a domain-specific encoder of $y$, $t$ is the sampling step, $\epsilon_\theta\left(z_t, t, \tau(y)\right)$ is the conditional denoising model with parameters $\theta$. 
% \jf{I remove the $\theta$ of $\tau_{\theta}$ since it is confusing for both $\epsilon$ and $\tau$ have subscript $\theta$.} 
The denoising process is modeled as a reverse process of a fixed Markov Chain of length $T$, and $t$ is uniformly sampled from $\{1, \ldots, T\}$. %$\epsilon_\theta\left(z_t, t, \tau_\theta(y)\right)$ is the conditional denoising autoencoder. $\theta$ represents the model parameters.

SD uses a frozen CLIP text encoder (i.e. $\tau$) to encode prompt words (i.e. $y$) and a UNet Model (i.e. $\epsilon_\theta$) for the denoising process~\citep{sd}. 
% \jf{Give a reference} 
Our experiments primarily rely on Stable Diffusion. The forward linear units in its UNet Model are fundamental to our proposed rank reduction.

%Our experiments primarily rely on the cutting-edge and widely-used Stable Diffusion, which is a variant of Latent Diffusion Models (LDMs) \citep{sd}.LDMs encode input images $x$ into a latent code $z$ with encoder $\mathcal{E}$, and then execute the denoising process on $z$. The training objective is 
%\begin{align} 
%\mathbb{E}_{\mathcal{E}(x), y, \epsilon \sim \mathcal{N}(0,1), t}\left[\left\|\epsilon-\epsilon_\theta\left(z_t, t, \tau_\theta(y)\right)\right\|_2^2\right]
%\end{align} 

%Here, $y$ is the input such as text, semantic maps. $\tau_\theta$ is a domain specific encoder of $y$. For the reverse process of a fixed Markov Chain of length $T$ for the denoising process, $t$ is uniformly sampled from $\{1, \ldots, T\}$. $\epsilon_\theta\left(z_t, t, \tau_\theta(y)\right)$ is the conditional denoising autoencoder. $\theta$ represents the model parameters.

\subsection{Fine-tuning Generative Models for Personalization}
Customizing and personalizing pre-trained text-to-image diffusion models has garnered significant research interest recently. Many methods have been proposed including enriching text embeddings~\citep{textinversion}, fine-tuning the UNet~\citep{dreambooth,kumari2023multi,hu2021lora,gandikota2023concept,han2023svdiff,lora_git}, 
% \jf{Previously, you use UNet instead of U-Net here. Better be consistent!} 
and providing adapters~\citep{mou2024t2i,zhang2023adding} to control the generated outcomes, as well as some training-free approaches~\citep{chen2024subject,gal2023designing,jia2023taming,shi2023instantbooth,wei2023elite}. 
% Methods like SVDiff \citep{han2023svdiff} have explored the spectral analysis of weights in attention mechanisms. However, their understanding is primarily limited to achieving parameter reduction, without recognizing that spectral filtering itself can serve as a method for feature constraint to enable customization.

Among the fine-tuning based personalized T2I methods, one common and effective idea is via matrix decomposition and adjustment of its components, which have been introduced in early GAN-based generative models~\citep{zhu2021low,feng2022rank}. 
% \jf{You need to give a space before the reference [44]. I have already corrected many similar issues.} 
The widely-used LoRA~\citep{hu2021lora,lora_git,gandikota2023concept} in diffusion models also follows the idea of matrix decomposition. Another method SVDiff~\citep{han2023svdiff} uses %more analytical approaches like 
SVD (singular value decomposition) to decompose matrices. However, these methods only focus on the scale of the obtained vector components. For example, LoRA ~\citep{lora_git,hu2021lora}
% \jf{Give a reference. There are too many LoRA applications. You need to specify which LoRA you are talking about.} 
adds the scale $\alpha$ to the original formula $W_0+BA$ as
\begin{align} \label{eq:LoRA}
W_0+{\alpha} \Delta W=W_0+{\alpha} BA, \text{where } W_0 \in \mathbb{R}^{d \times k}, B \in \mathbb{R}^{d \times r}, A \in \mathbb{R}^{r \times k}
\end{align} 
where $W_0$ is matrix of a pretrained SD model. 
% \jf{I replace $\frac{\alpha}{r}$ with $\alpha$ for simplicity since $r$ is a constant and to echo with Section 3.2.}  
Note that since the column vectors of $\Delta W$ are not normalized, $B$ and $A$ inherently include the learning of scales for the components. Similarly, SVDiff~\citep{han2023svdiff} optimizes the scales of the diagonal matrix obtained from the SVD decomposition of $W_0$. 

Our PaRa method, in contrast, chooses to directly eliminate certain components during personalization rather than adjusting their scales. Because neural networks often tend to overfit, eliminating some components can be more stable and robust than adjusting their scales~\citep{liu2015sparse,guo2016dynamic,han2015learning}. Furthermore, eliminating components is an idempotent operation (meaning it can be applied multiple times without changing the result to go beyond the initial application), making model mixing or combination more stable, compared to scale adjustments.

\subsection{Personalization-based Image Editing in Diffusion Models} 
% \jf{2.1 and 2.2, you only capitalize the first word but here you capitalize every word. Be consistent!}

Existing fine-tuning models \citep{dreambooth,kumari2023multi,hu2021lora,gandikota2023concept,han2023svdiff,zhang2023sine} that attempt to perform image editing directly by "one-shot training and adjusting the text of the original training image" are prone to overfitting to the single training image. Even if they can avoid overfitting, the diversity of images generated from the same text and Gaussian noise makes it nearly impossible to reproduce the exact training images during generation. Therefore, personalized diffusion models require the inversion process \citep{song2020denoising,mokady2023null}  to lock in the noise during image editing. SVDiff \citep{han2023svdiff} demonstrates that it can treat the inversion process as an optional component. However, omitting the inversion process still significantly impacts its faithfulness to the target image. In contrast, our model PaRa can maintain editability after only one-shot learning, and eliminate the need for the inversion process to achieve image editing.

%% file: sections/method.tex
\section{Method}

% \zizheng{it would be better to give an overview of this section. \eg, first, we xx, then, we introduce ..., finally, we xxx}

This section introduces the details of our PaRa model, including three sections.
Firstly, we explain the fundamental principles and implementation of PaRa, including its formulation, training process, and application to convolutional layers.
Secondly, we discuss how to combine two trained PaRa models, %. This part also explores 
what the relationship is between PaRa and LoRA, and how PaRa can be effectively used in conjunction with pre-trained LoRA models.
Lastly, we delve into the application of PaRa in single-image editing.

\label{sec:PaRa_approach}
\subsection{Parameter Rank Reduction}
\label{sec:PaRa_approach_base}
In this section, we describe our approach PaRa with a simple demonstration based on a linear projection.
Let $W_0 \in \mathbb{R}^{d \times k}$ be the weight matrix of one linear projection in a diffusion model, where $d$ and $k$ denote the number of input and output hidden dimensions, respectively. Given the input of this layer $x$, the output can be written as $h=W_0x$. Note that we omit the bias term for simplicity.
% For neurons in a neural network that perform matrix multiplication, given a pre-trained weight matrix $W_0 \in \mathbb{R}^{d \times k}$, 
% the layer output is computed as $h=W_0x$.
% LoRA updates the model by adding a low-rank matrix, expressed as $h=W_0x+\Delta Wx=W_0x+BAx$, where $B \in \mathbb{R}^{d \times r}, A \in \mathbb{R}^{r \times k}$.
To reduce the output space (\ie, column space) of $W_0$, we introduce a learnable parameter $B \in \mathbb{R}^{d \times r}$, where $r$ is the hyperparameter that controls the matrix rank.
%Our main idea is to reduce the output space (\ie, column space) of $W_0$ is to firstly 
We first decompose $B$ using QR decomposition
$B=QR$, where $Q$ is an orthogonal matrix and $R$ is the corresponding upper triangular matrix. Based on $Q$, we then compute $W_{reduce}$ as
\begin{align} \label{eq:PaRa}
W_{reduce} = W_0 - QQ^TW_0.
\end{align} 
Next, the rank-reduced layer outputs can be formulated as
\begin{align} \label{eq:PaRa_modi}
h = W_{reduce} x = W_0 x - QQ^T W_0 x.
\end{align} 
With a theoretical proof provided in Appendix \ref{sec:appendixA}, Eq.~\ref{eq:PaRa_modi} ensures that the column space of $W_{reduce}$ is a subset of the column space of $W_0$, effectively reducing the dimension of the output while maintaining the key features learned by the model.
% . For the reason, please refer to the theorem and proof in Appendix A. 
% In linear algebra, the output space is often referred to as the image space, but to avoid confusion with the pixel space, we will use the term output space.

In other words, given $W_0 = \begin{bmatrix}
\vec{w_{1}} \ \vec{w_{2}} \
...\ \vec{w_{k}},
\end{bmatrix}_{d\times k}$, where $\vec{w_{i}}$ denotes the column vector of $W_0$, 
% \zizheng{where $\vec{w_{i}}$ denotes xxx}
we define the column space of $W_0$ as  $S_0=\Span\{\vec{w_{1}},\vec{w_{2}},
... \vec{w_{k}}\}$. By adjusting $W_0$ to $W_{reduce}$, we ensure that the column space of $W_{reduce}$ is a subset of the column space of $W_0$.
In practice, we initialize $B$ to zero and finetune it with a few text-image pairs as the common practice in~\cite{dreambooth}. Our goal is to evolve $B$ into a set of orthonormal bases, and then adjust $W_0$ by subtracting its components on these bases. This process ensures the column space is effectively reduced.

Moreover, note that $QQ^T W_0$ now represents the components of $W_0$ projected onto the column space of $B$, capturing the influence of the orthonormal basis derived from  $B$ on $W_0$. When $B$ has linearly independent columns, the reduced dimension will be the same as $r$, i.e. the column number of $B$. Since $r$ is typically set to a small number ( \eg, 2 or 4, compared to hundreds/thousands of hidden dimensions in a layer output), $B$ is likely to have linearly independent columns after training.

\paragraph{Remark on convolutional kernels.}
% For the weights of the linear layers, we can directly use Equation \ref{eq:PaRa}. 
For the weights of convolutional layers, we need to employ a reshaping method similar to %that used in 
FSGAN~\citep{robb2020few} before reducing the rank. Specifically, we reshape the convolution kernel weight
$W_{0\_conv} \in \mathbb{R}^{c_{out} \times c_{in} \times h \times w}$ to the matrix as a second-order tensor
$W_0 \in \mathbb{R}^{c_{out} \times (c_{in} \times h \times w)}$. 
After this, we can proceed with the steps of PaRa, setting $B \in \mathbb{R}^{c_{out}\times (c_{in} \times h \times w)}$, $B=QR$, calculate and reshape $QQ^TW_0$ from ${c_{out}\times (c_{in} \times h \times w)}$ to ${c_{out}\times c_{in} \times h \times w}$ as $\Delta W$. Finally, the rank-reduced kernel weight becomes 
\begin{align} \label{eq:PaRa_conv}
    W_{reduce\_conv} = W_{0\_conv} - \Delta W,
\end{align}
where $ \Delta W \in \mathbb{R}^{c_{out} \times c_{in} \times h \times w}$.

\subsection{Combining PaRa}
\label{sec:CombiningPaRa}
Our framework also supports the combinations of different personalized PaRa models as well as combining with LoRA-based personalized T2I models. %the individually trained PaRa can be combined into a new model. 
Let's first consider two individually trained PaRa models 
\begin{align} \label{eq:two_PaRa}
    W_1=W_0-Q_1Q_1^TW_0,  W_2=W_0-Q_2Q_2^TW_0,
\end{align}
where
$Q1 = \begin{bmatrix} \vec{q_{11}} \ 
\vec{q_{12}} \  ... \vec{q_{1r_{1}}}
\end{bmatrix}_{d\times r_{1}}$ and
$Q2 = \begin{bmatrix} \vec{q_{21}} \ 
\vec{q_{22}} \  ... \vec{q_{2r_{2}}}
\end{bmatrix}_{d\times r_{2}}$. We can merge $Q_1$ and $Q_2$ into $
Qm = \begin{bmatrix} \vec{q_{11}} \ 
\vec{q_{12}} \  ... \vec{q_{1r_{1}}} \vec{q_{21}} \ 
\vec{q_{22}} \  ... \vec{q_{2r_{2}}}
\end{bmatrix}_{d\times (r_{1}+r_{2})}$ and reduce to a new orthonormal matrix $Q'_m$ by QR decomposition $Q_m=Q_m'R_m'$. Then the combined PaRa model has new weights $W_m = W_0-Q_m'Q_m'^TW_0$. 
In practice, a more convenient approach is via sequential addition to a diffusion model, where the first PaRa diffusion model is used as the new base diffusion model for the second one. %, and then another PaRa is directly added. 
This can be expressed as:
% When both $r_1$ and $r_2$ are sufficiently small, in practice, a more convenient approach is to use one PaRa diffusion model as the new base diffusion model, and then directly add another PaRa. This can be expressed as:
\begin{align} \label{eq:qrr_comb}
    W_m = W_1 - Q_2Q_2^TW_1 = (W_0-Q_1Q_1^TW_0)-Q_2Q_2^T (W_0 
 - Q_1Q_1^TW_0).
\end{align}
We provide a proof in the Appendix \ref{sec:appendixB} that this is equivalent to $W_m = W_0-Q_m'Q_m'^TW_0$. 
% However, this can only serve as an approximation when $r \ll k$ for $W_0$ with rank $k$, otherwise this method is not applicable, as explained in Appendix C.

\paragraph{Rank boundary.} 
% Not only in PaRa combination, but also in other contexts, we require $r \ll k$.
For a matrix $W_0$ with rank $k$, if the reduced rank $r$ is too large compared to $k$, it can cause the original model to collapse and fail to generate images properly. 
% In individual PaRa models, we avoid setting the the value of $r$ too close to $k$. However, in the process of combining PaRa models, 
Especially in the PaRa combination, we are uncertain about the exact value of the combined rank $r_{combine}$, for which we only know that it is greater than
% that $r_{combine}$ is greater than 
$r_1$ and $r_2$, but less than $r_1+r_2$. Thus, there is a possibility that 
$r_{combine}$ could be large enough to cause the original model to collapse.
Therefore, for an PaRa, even though we set the same $r$ for the entire pre-trained model, for each layer's weight $W_0$ in the pre-trained diffusion model, we impose an upper limit on the rank $r$ that can be reduced, denoted as  $r_{adjust} \leq \gamma rank(W_0)$, where $\gamma$ is a factor less than 1. For each $W_0$, the reduced rank is 
\begin{align} 
r_{adjust} = 
\begin{cases} 
r & \text{if } r \leq \gamma rank(W_0)\\
\lfloor \gamma rank(W_0) \rfloor & \text{if } r > \gamma rank(W_0)
\end{cases}
\end{align}

% \subsection{Comparison and Combination with LoRA}
\paragraph{Comparison and combination with LoRA.}
\label{LoRACombinesubsection}
PaRa has a corresponding relationship with the formulation of LoRA~\citep{lora_git}, $W_0 + \alpha BA$, 
% \jf{Give the reference to specify which LoRA.}
% Compared with the formulation of LoRA~\citep{lora_git}: $W_0+\alpha BA$, \jf{Give the reference to specify which LoRA.} PaRa has a corresponding relationship with it, 
i.e., we can consider $-Q$ as $B$ and $Q^{T}W_0$ as $A$. This means that we can %use the method to combine LoRA to 
combine individually trained LoRA and PaRa as $\Delta W=(-\alpha_1Q+\alpha_2B)(\alpha_1Q^TW_0+\alpha_2A)$, where $\alpha_1$ and $\alpha_2$ are parameters controlling the strength of PaRa and LoRA. 
However, this combination method treats PaRa entirely as LoRA, failing to preserve control over the diversity of the generated images.

According to the rank-reduction property of PaRa, it does not require the scaling parameter. In other words, PaRa effectively determines the optimal scale $\alpha$, under which it reduces the rank. Therefore, we propose to combine the models as 
$W_{combine} = W_0-QQ^TW_0+\alpha_{LoRA} BA$.
$\alpha_{LoRA}$ is the scale parameter of LoRA. In this way, we first reduce the rank and then add $BA$. 

A question naturally arises that whether it is feasible to first apply LoRA and then reduce the rank of the new weights, \ie, $W_{combine} = W_0+\alpha_{LoRA} BA-QQ^T(W_0+\alpha_{LoRA} BA)$. 
% The answer is no. 
This is technically feasible but does not work well in practice since the variations introduced by LoRA extend the activation space of the initial layer, \eg, learning a new concept on top of the existing data distribution. Placing PaRa afterwards would lead PaRa to attempt to eliminate these variations.
The larger the intersection between the column vector spaces of PaRa's $Q$ and LoRA's $B$, the more the effect of LoRA will be diminished. An extreme example is that if $Q$ is the same as the orthogonal basis of $B$, then regardless of how $\alpha_{LoRA}$ is adjusted, the LoRA model will not have any effect. Comparative experiments can be found in Section \ref{loracombineexp}.

\subsection{Performing Single Image Editing Like Text-to-Image Generation}
\label{sec:Imageedit}

% For most fine-tuning methods, image editing cannot be directly achieved by adjusting the text of the original training images. This is because the diversity of images generated from the same text and Gaussian noise makes it impossible to reproduce the exact training images during generation. Therefore, they require the inversion process \citep{song2020denoising,mokady2023null} to lock in the noise during image editing.

% PaRa (Projection-based Rank Reduction) 
PaRa can perform single image editing directly through one-shot training and adjusting the text of the original training image. 
PaRa offers a solution by stabilizing the output, such that different Gaussian noises tend to yield the same result. This stability means that the model can generate images that closely resemble the training image, even when using various text prompts. This enables direct modification of the text prompt to facilitate image editing on the single training image.

By controlling the rank in PaRa, we can balance between faithful reconstruction and editability. When a large rank is selected, the model produces images that are very similar to the training image, enhancing reconstruction fidelity. Conversely, selecting a smaller rank increases the diversity of the generated images, improving editability.
We provide a detailed mathematical discussion of how PaRa achieves this balance and the role of linear algebra concepts in Appendix \ref{sec:appendixG}.

%% file: sections/experiments.tex
\section{Experiments}
\label{sec:PaRa_exp}

The experiments evaluate PaRa on various tasks including single/multi-subject generation and single image editing, together with ablation studies. The SDXL1.0~\citep{sdxl} and the DDIM sampler with $\eta = 0$ are used for image generation. All experiments are conducted on a single A100 GPU with 40GB of VRAM.

\subsection{Single-Subject Generation}
% In this section, we compare the effects of PaRa with Dreambooth, Textual Inversion, and LoRA on customized single-subject generation, based on the Dreambooth dataset. All baselines were trained for 1000 steps with batch size 1, and the LoRA choosed rank $r=16$ and scale $\alpha=2.2$ as the best model for fair comparison. 
\textbf{Implementation details.} We evaluated the effects of PaRa on customized single-subject generation, based on the Dreambooth dataset. First, we verified that PaRa can indeed reduce the output space by confirming that the output diversity of PaRa has indeed decreased, compared to Vanilla SDXL1.0.  Then we compared the effects of PaRa at different ranks with baseline methods on customized single-subject generation, including Dreambooth~\cite{dreambooth}, Textual Inversion~\cite{textinversion}, SVDiff~\cite{han2023svdiff}, and LoRA~\cite{hu2021lora,lora_git}.  All baselines were trained for 1000 steps with a batch size of 1, and LoRA chose a rank of $r=16$ and a scale of $\alpha=2.2$ as the best model for a fair comparison. In our experiments, we found that PaRa already achieved ideal results at around 200 steps. Therefore, %as a challenge, 
we compare the results of PaRa at 200 steps with the baselines at 1000 steps. Also, we employ a rank boundary $\gamma = 1/40$ for PaRa.

\textbf{Evaluation metrics.}
We quantify the generation diversity using the metric of average SSIM. SSIM \citep{ssim} is commonly used to measure image similarity. For a generated image set, we calculate the SSIM for each pair of images and then compute the average value. 
We use the CLIP score $cos(\tilde{\mathbf{x}}, c)$ to measure the text alignment between the generated image and the text~\citep{radford2021learning}. Additionally, we compute $1-\mathcal{L}_{\text {LPIPS }}\left(\tilde{\mathbf{x}}\right)$ to measure the image alignment between the generated image and the training image~\citep{zhang2018unreasonable}. Here, the generated and training images are denoted as $\tilde{\mathbf{x}}$ and $\mathbf{x}$, respectively, and the prompt is denoted as $c$. %,  and the training image is denoted as $\mathbf{x}$.

\paragraph{Comparison of output diversity.}
\label{Output_Diversity}
To verify that the diversity of the output has indeed decreased in PaRa, we compared the average SSIM %of the generated results between 
of SDXL 1.0~\citep{sdxl} and its finetuned version using PaRa.  
% We compared the average SSIM of 500 generated images for various prompts under different random seeds in several categories of the Dreambooth dataset,
Based on 8 categories in Dreambooth, we generated 500 images for various prompts under different random seeds. Error bar is calculated by 5 runs.
% and repeated this experiment 5 times to obtain the average and error range, as shown in Table \ref{tab:SSIM}. 
% It can be observed that the SSIM of PaRa is significantly higher than that of vanilla SDXL.
As shown in Table \ref{tab:SSIM}, on single-subject generation, PaRa demonstrates higher SSIM scores, indicating less diverse generations, compared to SDXL.
% the average SSIM of PaRa is slightly higher compared to vanilla SDXL. This difference is small in single-subject generation due to the higher degree of freedom in the generated images. 
Furthermore, as shown in Fig. \ref{fig:imageEditPaRa}, on single-image editing, this advantage becomes more apparent since it achieves even better average SSIM scores. Overall, it supports our assumption that the image space of the linear transformation $W_0 + \Delta W$
becomes smaller than that of $W_0$.

\begin{table}[t]
% \vskip -0.1in
% // a headshot of a person $<$Male$>$ $<$Skin tone 5$>$ $<$Age 60 69$>$}
% \vskip -0.2in

\begin{center}
\begin{small}
\begin{sc}
\caption{Average SSIM results of two different prompts (top and bottom) with different subjects (each column) to indicate the diversity of the generated images. %, where each table corresponds to different experimental prompts, and each column represents a different subject. 
We compared Vanilla SDXL ~\citep{sdxl} with our PaRa using ranks 4 and 8. Higher average SSIM values suggest lower diversity and better alignment with the train images. We \textbf{bold} the results of the highest average SSIM values.}
\label{tab:SSIM}
\resizebox{1.0\linewidth}{!}{
\begin{tabular}{@{}ccccccccc@{}} 
\toprule
``a [V] sculpture made of gold'' \ \ \ \  \ \ \ \ \ \ \ \ \ \ \ \ \   & bear\_plushie & cat & dog8 & ducktoy & grey\_sloth\_plushie & monster\_toy & red\_cartoon & wolf\_plushie \\
% & \multicolumn{3}{c}{cat}  &&  \multicolumn{3}{c}{dog} &&  \multicolumn{3}{c}{ducktoy} \\
\midrule
Vanilla SDXL & 0.185$\pm$0.015&0.372$\pm$0.018&0.369$\pm$0.021&0.391$\pm$0.021&0.375$\pm$0.042&0.341$\pm$0.017&0.258$\pm$0.029&0.162$\pm$0.013\\
PaRa $r=4$&0.203$\pm$0.008&0.390$\pm$0.033&0.463$\pm$0.011&0.516$\pm$0.010&\textbf{0.382}$\pm$0.013&0.366$\pm$0.013&0.327$\pm$0.014&\textbf{0.170}$\pm$0.012\\
PaRa $r=8$& \textbf{0.237}$\pm$0.010&\textbf{0.401}$\pm$0.028&\textbf{0.465}$\pm$0.020&\textbf{0.519}$\pm$0.015&0.379$\pm$0.011&\textbf{0.381}$\pm$0.015&\textbf{0.386}$\pm$0.015&0.160$\pm$0.017\\
\bottomrule
\end{tabular}
}

\resizebox{1.0\linewidth}{!}{
\begin{tabular}{@{}ccccccccc@{}} 
\toprule
``A [V] on a skateboard in times square'' & bear\_plushie & cat & dog8 & ducktoy & grey\_sloth\_plushie & monster\_toy & red\_cartoon & wolf\_plushie \\
% & \multicolumn{3}{c}{cat}  &&  \multicolumn{3}{c}{dog} &&  \multicolumn{3}{c}{ducktoy} \\
\midrule
Vanilla SDXL & 0.294$\pm$0.029&0.354$\pm$0.031&0.305$\pm$0.038&0.326$\pm$0.078&0.287$\pm$0.013&0.262$\pm$0.010&0.233$\pm$0.066&0.201$\pm$0.014\\
PaRa $r=4$& 0.327$\pm$0.025&0.403$\pm$0.082&0.307$\pm$0.025&0.351$\pm$0.025&0.374$\pm$0.023&\textbf{0.276}$\pm$0.017&0.250$\pm$0.031&0.205$\pm$0.014\\
PaRa $r=8$& \textbf{0.339}$\pm$0.036&\textbf{0.405}$\pm$0.110&\textbf{0.323}$\pm$0.016&\textbf{0.359}$\pm$0.031&\textbf{0.383}$\pm$0.012&0.272$\pm$0.014&\textbf{0.251}$\pm$0.044&\textbf{0.232}$\pm$0.013\\
\bottomrule
\end{tabular}
}

% \resizebox{1.0\linewidth}{!}{
% \begin{tabular}{@{}ccccccccc@{}} 
% \toprule
% ``A [V] in a swimming pool'' \ \ \ \  \ \ \ \ \ \ \ \ \ \ \ \ \ \ \ \ \ \ \ \ \ \   & bear\_plushie & cat & dog8 & ducktoy & grey\_sloth\_plushie & monster\_toy & red\_cartoon & wolf\_plushie \\
% % & \multicolumn{3}{c}{cat}  &&  \multicolumn{3}{c}{dog} &&  \multicolumn{3}{c}{ducktoy} \\
% \midrule
% Vanilla SDXL & ?$\pm$?&?$\pm$?&0.435$\pm$?&?$\pm$?&?$\pm$?&?$\pm$?&?$\pm$?&?$\pm$?\\
% PaRa $r=4$&0.191$\pm$0.003&0.501$\pm$0.004&0.378$\pm$0.002&0.550$\pm$0.013&0.379$\pm$0.005&0.267$\pm$0.002&0.448$\pm$0.031&0.489$\pm$0.011\\
% PaRa $r=8$& 0.194$\pm$0.003&0.499$\pm$0.006&0.353$\pm$0.003&0.403$\pm$0.003&0.332$\pm$0.007&0.422$\pm$0.026&0.417$\pm$0.008&0.422$\pm$0.004\\
% \bottomrule
% \end{tabular}
% }
\end{sc}
\end{small}
\end{center}
\vskip -0.2in
\end{table}

\begin{figure}
% \vskip -0.1in
  \centering
\centerline{\includegraphics[width=0.99\columnwidth]{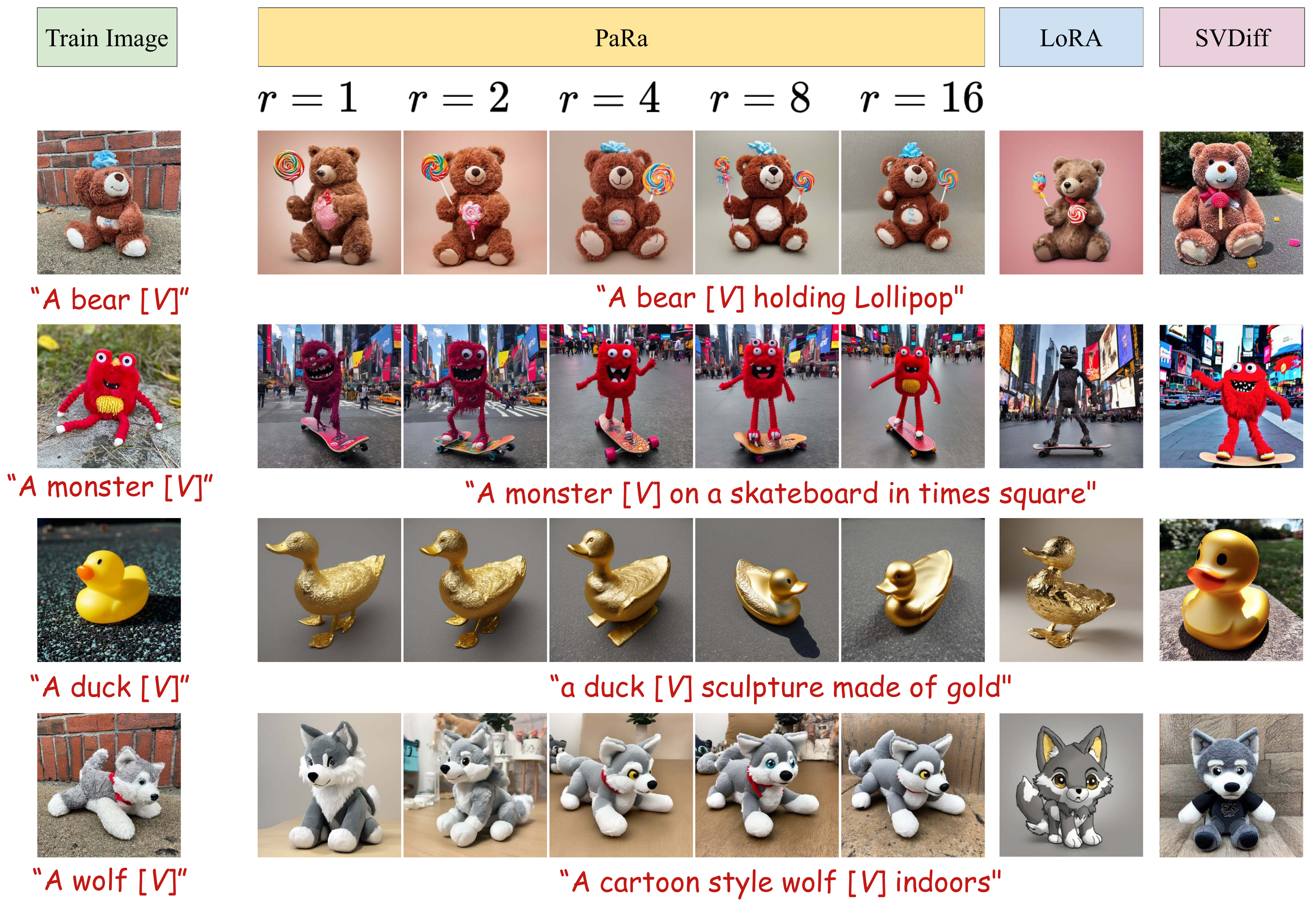}}
  \caption{Comparing the proposed PaRa with LoRA~\cite{hu2021lora} and SVDiff~\cite{han2023svdiff} on single-subject generation. Each subject has 5 training images. PaRa includes results with ranks $r$ ranging from 1 to 16. For LoRA, we adopt a rank of 8.
  % where we can see that the larger the rank $r$, the closer the generated images are to the training images. 
  % We set the rank of LoRA to 8.
  % from our experiments with empirically selected 
  % scale values of 1.0 and 2.2.
  % Clearly, in terms of alignment with the training images, PaRa is better than LoRA and SVDiff. 
 We provide more generation results on Dreambooth and Textual Inversion in Appendix \ref{sec:appendixF}.}
\label{singleimagegenerate}
\vskip -0.1in
\end{figure}

\begin{figure}[ht]

\centering            \includegraphics[width=\linewidth]{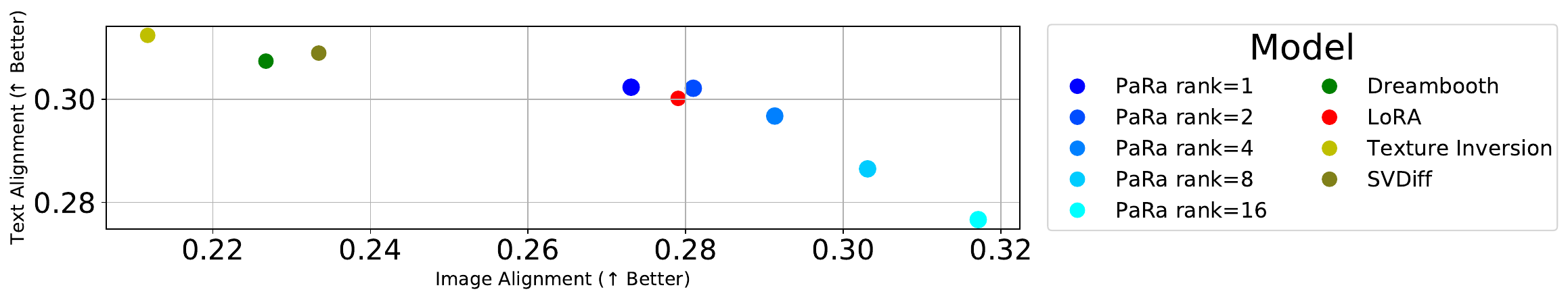}

\caption{Text and image alignments for single subject generations. Text Alignment is measured by the CLIP score $cos(\tilde{\mathbf{x}}, c)$, and Image Alignmnet is measured by $1-\mathcal{L}_{\text {LPIPS }}\left(\tilde{\mathbf{x}}\right)$.
% In figure (a), our model PaRa (green points)is positioned towards the lower left corner compared to other models. In figure (b), as the rank increases, the benchmark extends towards the lower left corner. 
The further to the right indicates higher fidelity to the personalized target, while higher along the vertical axis indicates improved text editability.}
% \vskip -0.1in
\label{fig:text_image_alignment}
\vskip -0.1in
\end{figure}

\paragraph{Comparison of generation quality.}
% To evaluate the quality of T2I model personalization, we compare the similarity between the generated results and the training images, as well as the alignment between the generated results and the prompts. 
In Fig. \ref{singleimagegenerate}, we compare PaRa with representative  personalization techniques for T2I models, \ie, LoRA~\cite{hu2021lora} and SVDiff~\cite{han2023svdiff}. Overall, PaRa consistently achieves better image alignment performance across different prompts under different ranks. Moreover, in Fig. \ref{fig:text_image_alignment}, we benchmark PaRa across all subjects and report the text alignment and image alignment scores under different ranks.
% We evaluated our model through a Human Reflection survey and a benchmark for text and image alignment. Details of this survey are provided in Appendix \ref{sec:appendixH}.
% Section \ref{ablation_section}.
% The benchmarks of the text and image alignment, averaged across , are presented in 
% The benchmarks of the text and image alignment are presented in Fig. \ref{fig:text_image_alignment}.  
% In Fig. \ref{fig:text_image_alignment} (a), the value of $r$ for PaRa has been set to a moderate value of 8.
It can be seen that PaRa results are positioned in the lower right part, indicating that PaRa achieves %generally better frontier and 
much better image alignment. Note that the relatively lower text-alignment ability of PaRa can be improved by reducing the rank $r$. %implies that PaRa is less sensitive to changes in the input prompt. Furthermore, we observe 
In general, a larger value of $r$ usually helps to align the generated image with the learned concept, while a smaller value of $r$ allows the text prompt to more effectively control the generated image. More results are provided in Appendix Fig.~\ref{fig:text_image_alignment_appendix}. 
 % Finally, we provide a visual comparison of single subject generation using PaRa with different values of $r$ and LoRA as a representative of other customization techniques in Fig. \ref{singleimagegenerate}.  

\subsection{Multi-Subject Generation}
\label{sec:MultiSubjectGeneration}
We demonstrate the effectiveness of combining PaRas which are individually trained on different subjects. 
Mixing two T2I model weights in previous works~\cite{wu2023mole,lora_git} generally result in problematic generation results, where they tend to over-emphasize one subject or mix two subjects to a `hybrid' entity that does not exist in reality. Moreover,
these methods require auxiliary techniques such as Cut-Mix-Unmix data augmentation ~\citep{han2023svdiff}, spatial condition ~\citep{gu2024mix}, where the additional data annotation and training bring non-negligible financial or computational costs.
In contrast, under our proposed PaRa, combining differently trained weights effectively eliminates the need for any auxiliary techniques while it can generate multiple subjects in a single image with little unrealistic blending.

% Here, we demonstrate that in combining PaRa models, without incorporating any auxiliary techniques, directly merging the models can effectively generate multiple subjects in a single image without unrealistic blending.

\begin{figure}
% \vskip -0.1in
  \centering
\centerline{\includegraphics[width=\columnwidth]{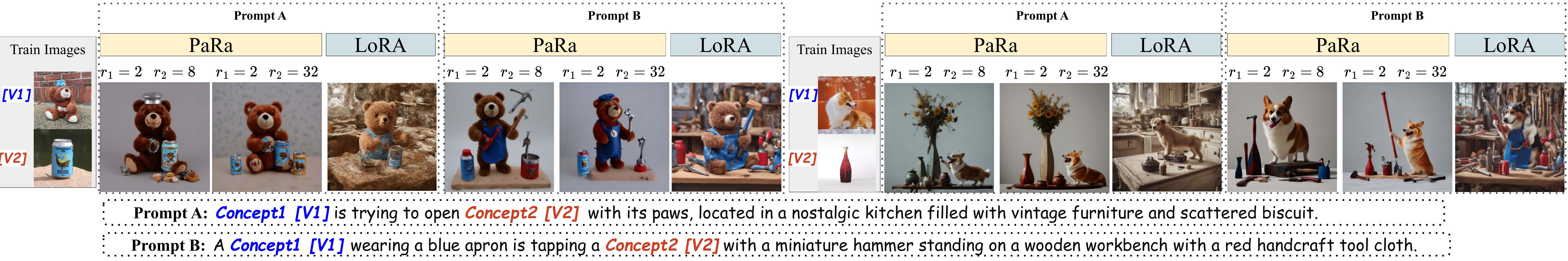}}
  \caption{Examples of multi-subject generation results. The PaRa results are generated with the reduced rank $r_1=2$ for Concept 1 and $r_2$ being 8 and 32 for Concept 2. LoRA results are generated with the best rank of 16 for both concepts and the scale value of 1.}
\label{fig:multigenerate}
\end{figure}

The results shown in Fig. \ref{fig:multigenerate} were obtained using two challenging prompts. It can be observed that LoRA~\cite{hu2021lora,lora_git} only manages to generate the primary concept, Concept1 [V1], into the generated images, while the second concept, Concept2 [V2], significantly deviates from the original image. In contrast, PaRa successfully captures both concepts well. 
% However, a drawback of PaRa is that when the reduced ranks 
% $r_1$ and $r_2$ of the two PaRa models are relatively large (as in the images in the bottom right corner of each group), it tends to generate simpler images.
As Cut-Mix-Unmix introduces unfair advantages due to the extra training data, we did not include it in our comparisons. More results can be found in Fig. \ref{fig:multigenerate_appendix} in Appendix \ref{sec:appendixE}.

\subsection{Single Image Editing}
\label{Single_Image_Editing}
\begin{figure}[ht]
% \vskip -0.1in
  \centering
\centerline{\includegraphics[width=\columnwidth]{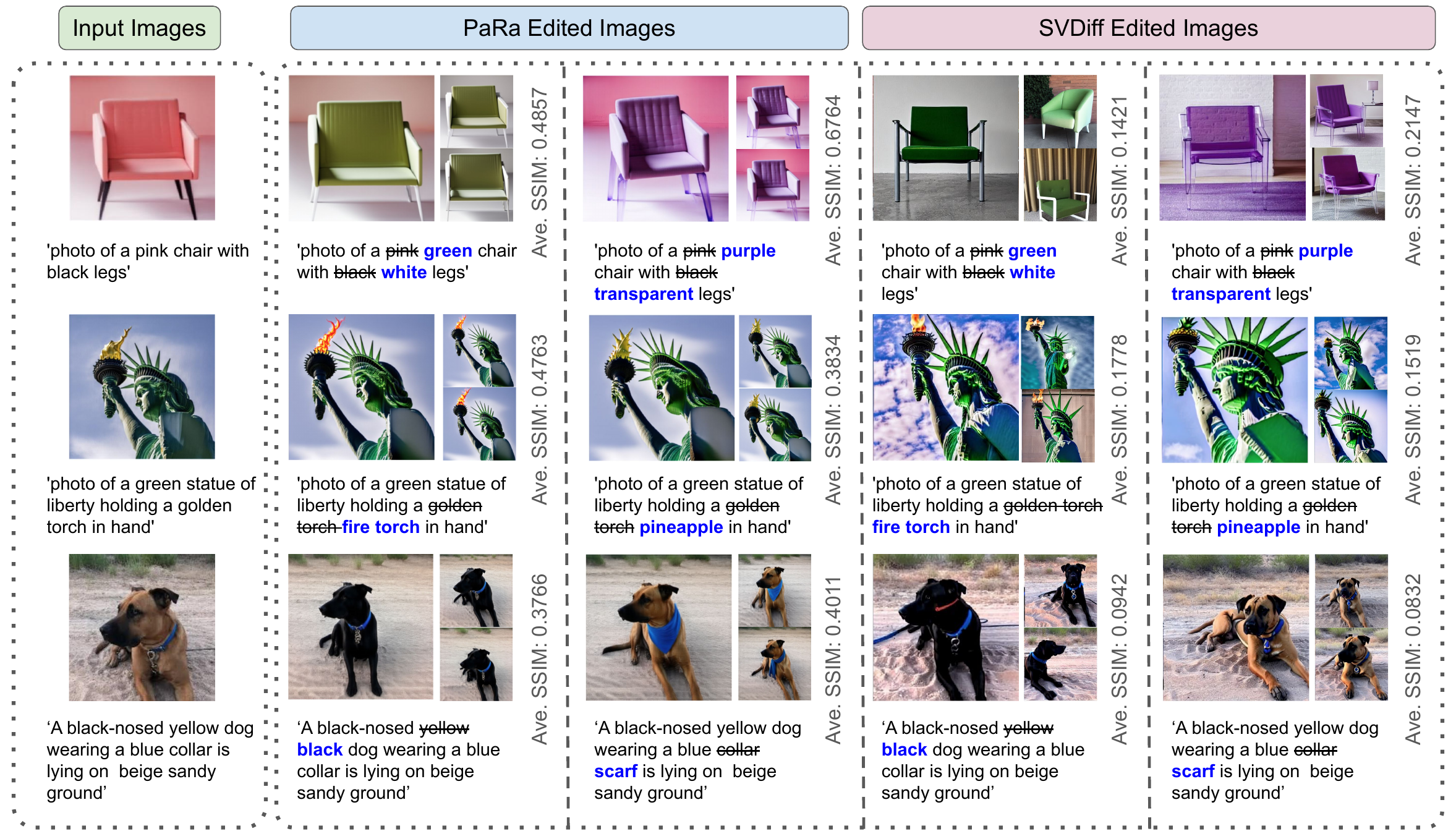}}
  \caption{Single-image editing results. PaRa allows for image editing through one-shot learning of the original image and performs generation by directly modifying the prompt. We can see that PaRa achieves the expected modifications and preserves the personalized target subject well. In addition, PaRa achieves high consistency with untargeted elements of the initial image under different Gaussian noises. %  , the generated images may slightly differ from the original, but the main subject remains well-preserved.
  }
\label{fig:imageEditPaRa}
% \vskip -0.1in
\end{figure}

In this section, we present the results of direct editing of a single image based on PaRa with $r=8$. The experiment in Fig. \ref{fig:imageEditPaRa} aims to demonstrate that with a properly chosen $r$, PaRa does not generate overly creative results and avoids the language drift issue~\citep{dreambooth}. As a comparison, we also present the results of SVDiff~\citep{han2023svdiff} on the same task without DDIM inversion. SVDiff can also partially achieve image editing effects without noise tracking (e.g. DDIM inversion). Here, we can see that PaRa has significantly less deviation from the original image %without noise tracking 
compared to SVDiff. Numerically, for each target image and prompt pair, we generated 100 images and calculated their average SSIM, where the result of PaRa was significantly higher, reflecting the stability of the generated images.

\subsection{Combination with LoRA}
\label{loracombineexp}
\begin{figure}[ht]
% \vskip -0.1in
  \centering
\centerline{\includegraphics[width=\columnwidth]{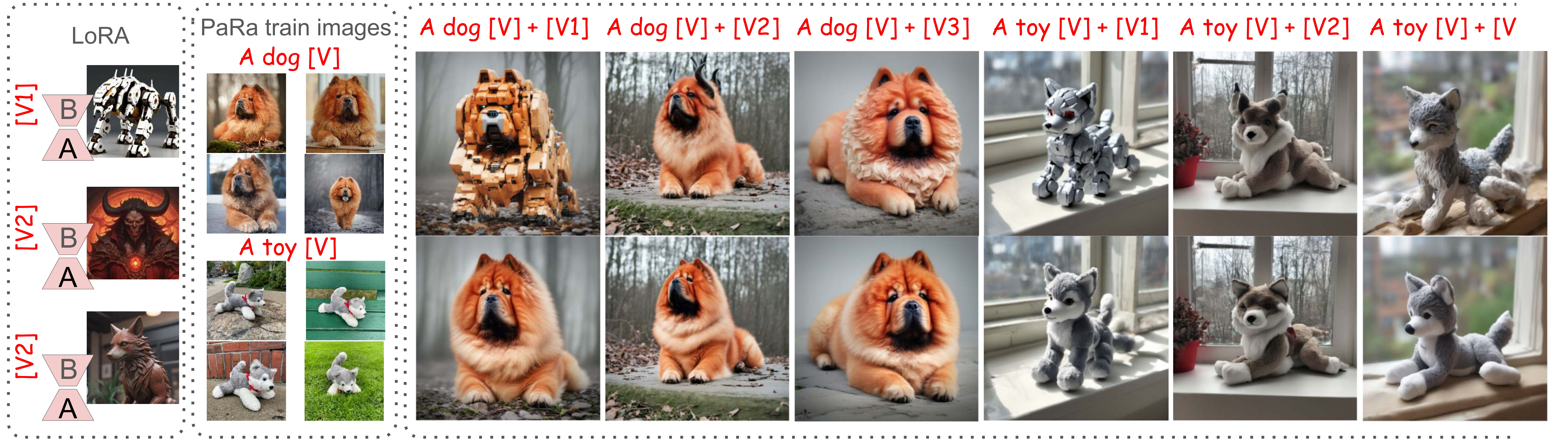}}
  \caption{ Generation results of combining PaRa with LoRA. Top: the results of first applying PaRa and then LoRA; Bottom: the results of first applying LoRA and then PaRa.}
\label{fig:PaRaLoRAcombine}
% \vskip -0.2in
\end{figure}

In Fig. \ref{fig:PaRaLoRAcombine}, we show the effects of combining PaRa and LoRA~\cite{hu2021lora,lora_git}. We adopt public pre-trained LoRA models \citep{mecha,figurine,diablo} as examples to demonstrate the compatibility with PaRa weights. In general, PaRa complements existing LoRA weights as it naturally combines the strength of both models, \ie, a learned novel concept and image generation style. 
% We note that during this combination, it is important to first apply PaRa and then LoRA. Applying LoRA first and then PaRa will result in the features of the LoRA model not being adequately represented.
Here, we demonstrate our discussion in Section \ref{LoRACombinesubsection}: during this combination, it is important to first apply PaRa and then LoRA. Conversely, applying LoRA first and then PaRa will, as shown in the figure, severely affect the performance of LoRA.

% \begin{figure}[ht]

% \centering            \includegraphics[width=1.0\linewidth]{figure/clipcossim_score_less_points.pdf}

% \caption{Text and image alignment for single subject generation. 
% In figure (a), our model PaRa (green points)is positioned towards the lower left corner compared to other models. In figure (b), as the rank increases, the benchmark extends towards the lower left corner. The further to the right indicates higher fidelity to the personalized target, while higher along the vertical axis indicates improved text editability.}
% % \vskip -0.1in
% \label{fig:text_image_alignment}
% \vskip -0.2in
% \end{figure}

\subsection{Ablation and Other Analysis}
\label{ablation_section}
% \paragraph{Parameter subsets and model sizes} We evaluated the different subset parameters of UNet in Appendix \ref{sec:appendixB}, which includes the model scale of PaRa within different parameter subsets and the generation performance for different subset parameters.

\textbf{Parameter subsets and model sizes.}   We evaluated using different subset parameters of UNet to update weights in Appendix \ref{sec:appendixC}, which includes the model sizes of PaRa within different parameter subsets for finetuning and the corresponding generation performance. % for different subset parameters.

\textbf{Rank bound $\gamma$.}  
In Section \ref{sec:CombiningPaRa}, we discuss the need for the rank boundary hyperparameter $\gamma$ to prevent the reduced rank from becoming too large, which could cause the generative model to collapse. In Section \ref{sec:MultiSubjectGeneration}, we mention that our chosen rank boundary of $\gamma = 1/40$ is based on empirical results. %it being the optimal choice from experimental results. 
In Appendix \ref{sec:appendixD}, we provide such empirical results by comparinging the effects of different $\gamma$.

% \paragraph{Comparing Different Rank Reductions 
% $r$}
\textbf{Comparing different rank reductions 
$r$.}  
The effect of using different rank reductions $r$ in PaRa is one of the central topics of this paper. As $r$ increases, the diversity of the generated images decreases, leading to more faithful reconstructions in image editing; as $r$ decreases, the diversity of the generated images approaches that of the original %SD and other fine-tuning 
models, improving text editability in image editing; if $r$ is too large, the generative model collapses. 
% We have already compared different values of $r$ in single-subject generation and multi-subject generation.
In Appendix \ref{sec:appendixE}, we compare the performance of different $r$ values in image editing, multi-subject generation, and LoRA combination.

% \paragraph{Human Reflection on PaRa Performance}
% \textbf{Human Reflection on PaRa Performance}  
% To evaluate the quality of the generated images, we conducted a survey where \SHYUPLAN{200} \SHYUPLAN{update later} participants provided their feedback. In order to streamline and ensure effectiveness, we designed a survey form consisting of nine questions. These questions allowed users to compare the image generation performance of PaRa and LoRA, covering both single-subject and multi-subject generation, as well as comparisons across different ranks. 
% % The overall results are summarized in Table \ref{tab:Survey}. 
% The link of the survey form and the statistical results are provided in the Appendix \ref{sec:appendixH}.

%% file: sections/conclusion.tex
\section{Conclusion}
\label{sec:PaRa_conclusion}
In conclusion, we have proposed PaRa, an effective and efficient framework for personalizing T2I diffusion models via parameter rank reduction. Our PaRa that finetunes with fewer parameters based on rank reduction can achieve better results than the prevailing LoRA fine-tuning. PaRa can well balance the diversity of generated images and the faithfulness to the customization objectives by selecting different ranks. PaRa is a flexible framework and performs well in different applications including single-subject generation, multi-subject generation, and single-image editing.
%Our method is a general approach that can also be applied to modalities such as text, video, and audio.

\textbf{Limitations.}   PaRa focuses only on reducing the original output space. However, we acknowledge that because current diffusion models are not yet universal models, sometimes customization might still need to expand the original output space. Although we assume that as the generative model grows larger, the need for expansion will decrease, a model that can simultaneously address both space extension and space reduction may be stronger. There are algebraic problems concerning space extension hidden here that might be worth exploring.

%% file: sections/appendix.tex
% \section{Appendix / supplemental material}
\label{sec:PaRa_appendix}

\appendix

\renewcommand \thepart{}

\renewcommand \partname{}

\part{Appendix}
We organize our supplementary material as follows.
\begin{itemize}
    \item In Section \ref{sec:appendixA}, we provide the theoretical basis for PaRa, including the proofs mentioned in Section \ref{sec:PaRa_approach_base}.
 
    \item In Section \ref{sec:appendixB}, we provide the proof mentioned in Section \ref{sec:CombiningPaRa}, explaining why PaRa Combination can be achieved through PaRa Sequential Addition.

    \item In Section \ref{sec:appendixC}, we provide a comparison of the model sizes and generated results when different subsets of SDXL parameters are updated using PaRa.
    % as mentioned in the experiments on parameter subsets and model sizes in Section \ref{ablation_section}.

    \item In Section \ref{sec:appendixD}, we compare the generating results of selecting different Rank Boundaries $\gamma$.

    \item In Section \ref{sec:appendixE}, we compare the performance of different $r$ values on the tasks of image editing and PaRa model combination.

    \item In Section \ref{sec:appendixF}, we present additional generating results of PaRa, including comparisons with DreamBooth~\citep{dreambooth} and Textual Inversion~\citep{textinversion} on single-subject generation. We also provide detailed images from Fig. \ref{fig:text_image_alignment} showing the performance on various concepts in the DreamBooth dataset, as well as more examples of effective PaRa combinations for multi-subject generation.

    \item In Section \ref{sec:appendixG}, we provide a mathematical argument demonstrating why directly generating with PaRa in one-shot learning is an effective method for image editing, as a supplement to Section \ref{sec:Imageedit}.

    \item In Section \ref{sec:appendixH}, we present the results of a survey conducted to evaluate the quality of the generated images, with 209 participants.

\end{itemize}

% \section{Appendix / supplemental material}
\section{Theorem and Proof of PaRa}
\label{sec:appendixA}
\begin{theorem}
  For matrix $W$, the image space of $W$ is a d-dimension vector space $S_d$. If we have matrix $Q=\begin{bmatrix}
\vec{q_{1}} \ \vec{q_{2}} \
...\ \vec{q_{r}}
\end{bmatrix}$ with $q_i \in S_d$ and vectors $q_i$ are mutually orthonormal, then $W-QQ^TW$ has a (d-r)-dimension image space.
\end{theorem}

This is actually an intuitive result, and to minimize any potential misleading, we provide a proof here.

% \textbf{Proof}

% The dimension of the image space of a matrix is same as the dimension of the column space of the matrix.

% For all column vectors of $Q$ and $W$, $\{\vec{q}_{1},\vec{q}_{2} ... \vec{q}_{r} , \vec{w}_i\} \in S_d$, we perform the Gram-Schmidt process to obtain an orthonormal basis $\{\vec{q}_{1}, \vec{q}_{2} ... \vec{q}_{r}, \vec{v}_1,\vec{v}_2 ... \vec{v}_{d-r}\}$ of the image space of $W$.

% $W-QQ^TW$ transforms each column vector $\vec{w_i}$ of $W$ into $\vec{w}_i-\sum_{j=1}^{r}(\vec{q}_j \cdot \vec{w}_i)\vec{q}_j$, which is a linear combination of  $\{\vec{v}_1,\vec{v}_2 ... \vec{v}_{d-r}\}$. Then $W-QQ^TW$ has a (d-r)-dimension image space.

\textbf{Proof}

Assume $W$ is a $b \times k$ matrix.
Denote $W-QQ^TW$ as 
$[\vec{u}_1 \ \vec{u}_2 ... \ \vec{u}_k]$. Denote $W$ as $[\vec{w}_1 \ \vec{w}_2 ... \ \vec{w}_k]$.
Present $W-QQ^TW$ in a vector form is

$[\vec{u}_1 \ \vec{u}_2 ... \ \vec{u}_k] = [\vec{w}_1 \ \vec{w}_2 ... \ \vec{w}_k] - [\vec{q}_1 \ \vec{q}_2 ... \ \vec{q}_r] \left[\begin{matrix}
\vec{q}_1^T \\
\vec{q}_2^T \\
... \\
\vec{q}_r^T \\
\end{matrix}\right][\vec{w}_1 \ \vec{w}_2 ... \ \vec{w}_k]  $

$=  \left[\vec{w}_1 \ \vec{w}_2 ... \ \vec{w}_k\right] - \left[\vec{q}_1 \ \vec{q}_2 ... \ \vec{q}_r \right]  
\left[\begin{matrix} 
\vec{q}_1^T \vec{w}_1&\vec{q}_1^T \vec{w}_2 &...&\vec{q}_1^T \vec{w}_k \\
\vec{q}_2^T \vec{w}_1&\vec{q}_2^T \vec{w}_2 &...&\vec{q}_2^T \vec{w}_k \\
... &...&...&...\\
\vec{q}_r^T \vec{w}_1&\vec{q}_r^T \vec{w}_2 &...&\vec{q}_r^T \vec{w}_k 
\end{matrix}\right]
$

$= \left[\vec{w}_1 \ \vec{w}_2 ... \ \vec{w}_k\right] - \left[\begin{matrix}
\sum_{i=1}^r \vec{q}_i \vec{q}_i^T \vec{w}_1 & \sum_{i=1}^r \vec{q}_i \vec{q}_i^T \vec{w}_2 &...& \sum_{i=1}^r \vec{q}_i \vec{q}_i^T \vec{w}_k
\end{matrix}\right]$

For each column, it is $\vec{u}_j = \vec{w}_j - \sum_{i=1}^r \vec{q}_i \vec{q}_i^T \vec{w}_j$

The image space of $W$ is a d-dimension vector space $S_d$, then we have $d < b$ and $d<k$.

($W-QQ^TW$) has a rank $(d-r)$ means we need to show $\exists \text{ independent } \{\vec{g}_1, \vec{g}_2, ... \vec{g}_{d-r}\}$, $\vec{g}_i \in S_d$, such that $\forall \vec{u}_i \in \{\vec{u}_1, \ \vec{u}_2 ... \ \vec{u}_k\}$, $\exists$ scalars $a_1, ... a_{d-r}\in\mathbb{R}$, \text{s.t. } $\vec{u}_i = \sum_{i=1}^{d-r}\vec{a}_i \vec{g}_i$.

 Construct a new set $P$ containing all column vectors of the matrices $Q$ and $W$, $P = \{\vec{q}_1, \vec{q}_2, ...,\vec{q}_r,\vec{w}_1, \vec{w}_2, ..., \vec{w}_k\}$. Apply the Gram-Schmidt process to $P$, we will get $d$ orthogonal basis vectors $\{\vec{v}_1, \vec{v}_2, ..., \vec{v}_d\}$

Apply the Gram-Schmidt process to $P$:

Step1: $\vec{v}_1 = \vec{q}_1$

Step2: $\vec{v}_2 = \vec{q}_2 - \frac{\vec{q}_1 \cdot \vec{q}_2}{\vec{q}_1 \cdot \vec{q}_1}  \vec{q}_1 = \vec{q}_2$

Step3: $\vec{v}_3 = \vec{q}_3 - \frac{\vec{q}_1 \cdot \vec{q}_3}{\vec{q}_1 \cdot \vec{q}_1}  \vec{q}_1  - \frac{\vec{q}_2 \cdot \vec{q}_3}{\vec{q}_2 \cdot \vec{q}_2}= \vec{q}_3$

...

Step(r): $\vec{v}_r = \vec{q}_r$

Step(r+1): $\vec{v}_{r+1} = \vec{w}_1 - \sum_{i=1}^{r} \frac{\vec{q}_i \cdot \vec{w}_1}{\vec{q}_i \cdot \vec{q}_i}  \vec{q}_i$

Step(r+2): $\vec{v}_{r+2} = \vec{w}_2 - \sum_{i=1}^{r} \frac{\vec{q}_i \cdot \vec{w}_2}{\vec{q}_i \cdot \vec{q}_i}  \vec{q}_i - \frac{\vec{v}_1 \cdot \vec{w}_2}{\vec{v}_1 \cdot \vec{v}_1}  \vec{v}_1$

...

As we only have $d$ basis vectors, we will have the orthogonal basis $\{\vec{q}_1, \vec{q}_2, ...,\vec{q}_r,\vec{v}_{r+1}, ..., \vec{v}_d\}$ finally.

Assume $\vec{w}_j = \sum_{l=1}^r a_l \vec{q}_l + \sum_{l=r+1}^d b_l \vec{v}_l $

$\vec{u}_j = \vec{w}_j - \sum_{i=1}^r \vec{q}_i \vec{q}_i^T \vec{w}_j \\
=\sum_{l=1}^r a_l \vec{q}_l + \sum_{l=r+1}^d b_l \vec{v}_l - \sum_{i=1}^r \vec{q}_i \vec{q}_i^T(\sum_{l=1}^r a_l \vec{q}_l + \sum_{l=r+1}^d b_l \vec{v}_l)\\
= \sum_{l=1}^r a_l \vec{q}_l + \sum_{l=r+1}^d b_l \vec{v}_l - \sum_{l=1}^r a_l \vec{q}_l \quad \text{(as $\vec{q}_i^T\vec{q}_l = 1$ if $i=l$, $\vec{q}_i^T\vec{q}_l = 0$ if $i \neq l$, $\vec{q}_i^T\vec{v}_l=0$ if $l \geq r+1$)}\\
= \sum_{l=r+1}^d b_l \vec{v}_l
$

As we want to proof $\exists \{\vec{g}_1, \vec{g}_2, ... \vec{g}_{d-r}\}$, \text{ s.t. } $\vec{g}_i \in S_d$, $\forall \vec{u}_i \in \{\vec{u}_1, \ \vec{u}_2 ... \ \vec{u}_k\}$, $\exists \{a_1, ... a_{d-r}\}$, \text{s.t. } $\vec{u}_i = \sum_{i=1}^{d-r}\vec{a}_i \vec{g}_i, a_i \in \mathbb{R}$,

so we have $\{\vec{g}_1, \vec{g}_2, ... \vec{g}_{d-r}\} = \{\vec{v}_{r+1}, ..., \vec{v}_d\}$, \text{ s.t. } $\vec{g}_i \in S_d$, $\forall \vec{u}_i \in \{\vec{u}_1, \ \vec{u}_2 ... \ \vec{u}_k\}$, $\exists \{a_1, ... a_{d-r}\}$, \text{s.t. } $\vec{u}_i = \sum_{i=1}^{d-r}\vec{a}_i \vec{g}_i, a_i \in \mathbb{R}$

Therefore, $W-QQ^TW$ has a (d-r)-dimension image space.

\textbf{Note:} The case not covered by this theorem is when there exists $Q$'s column vector $\vec{q}_l \notin S_d$. When $\vec{q}_l \notin S_d$, we have $\vec{q}_l \cdot \vec{w}_j=0$ for any column vector $\vec{w}_j$ of $W$. That is, this component $\vec{q}_l$ will not update $W$ in $W-QQ^TW$, because $\vec{w}_j - \vec{q}_l \vec{q}_l^T \vec{w}_j = 0$. In PaRa, if the PaRa training has not yet converged, then this $\vec{q}_l$ will be updated. If it has already converged, it indicates that our choice of $r$ might be too large, leading to over-parameterization of PaRa, but the other $\vec{q}_i \in S_d$ can still achieve the objective of personalization.

\section{Proof of PaRa Sequential Addition in PaRa Combination}
\label{sec:appendixB}

Here we will prove that in PaRa combination, $W_0-Q_m'Q_m'^TW_0 = W_1 - Q_2Q_2^TW_1 = W_0-Q_1Q_1^TW_0-Q_2Q_2^TW_0+Q_2Q_2^TQ_1Q_1^TW_0$.

For two trained PaRa models, we have parameters:

$Q1 = \begin{bmatrix} \vec{q}_{11} \ 
\vec{q}_{12} \  ... \vec{q}_{1r_{1}}
\end{bmatrix}_{d\times r_{1}}$ 

$Q2 = \begin{bmatrix} \vec{q}_{21} \ 
\vec{q}_{22} \  ... \vec{q}_{2r_{2}}
\end{bmatrix}_{d\times r_{2}}$. 

For the standard combination of  $Q_1$ and $Q_2$: $
Qm = \begin{bmatrix} \vec{q}_{11} \ 
\vec{q}_{12} \  ... \vec{q}_{1r_{1}} \vec{q}_{21} \ 
\vec{q}_{22} \  ... \vec{q}_{2r_{2}}
\end{bmatrix}_{d\times (r_{1}+r_{2})}$

With $Q_m'R_m'=Q_m$, same as what we dicussed in Appendix A, we can find out a orthonormal basis $\{\vec{q}_{11}, \vec{q}_{12} ... \vec{q}_{1r_1}, \vec{p}_1,\vec{p}_2 ... \vec{p}_{(r_m-r_1)}\}$ of $r_m$ vectors from the column vectors of $Q_1$ and $Q_2$.

$W_m = W_0-Q_m'Q_m'^TW_0$ means for each column vector $\vec{w_i}$ of $W$, it is transferred to 
\begin{align} 
\vec{w_i} - \sum_{j}^{r_1}(\vec{q}_{1j} \cdot \vec{w}_i)\vec{q}_{1j} - \sum_{l}^{r_m - r_1}(\vec{p}_{l} \cdot \vec{w}_i)\vec{p}_{l}
\label{Ex:combineTheory}
\end{align} 

For the practical combination
$W_m = W_1 - Q_2Q_2^TW_1 = W_0-Q_1Q_1^TW_0-Q_2Q_2^TW_0+Q_2Q_2^TQ_1Q_1^TW_0$, $\vec{w_i}$ is transferred to
\begin{align} 
\vec{w_i} - \sum_{j}^{r_1}(\vec{q}_{1j} \cdot \vec{w}_i)\vec{q}_{1j} - \sum_{s}^{r_2}(\vec{q}_{2s} \cdot \vec{w}_i)\vec{q}_{2s} +
\sum_{s}^{r_2}((\sum_{j}^{r_1}(\vec{q}_{1j} \cdot \vec{w}_i)\vec{q}_{1j})\cdot \vec{q}_{2s})\vec{q}_{2s}
\label{Ex:combinePrac}
\end{align} 

% We have $\vec{q}_{2s}=\sum_{j}^{r_1}(\vec{q}_{1j} \cdot \vec{q}_{2s})\vec{q}_{1j}+
% \sum_{l}^{r_m - r_1}(\vec{p}_{l} \cdot \vec{q}_{2s})\vec{p}_{l}$, then expression \ref{Ex:combinePrac} is 

We have $\vec{q}_{2s}=\sum_{j}^{r_1}a_{sj}\vec{q}_{1j}+
\sum_{l}^{r_m - r_1}b_{sl}\vec{p}_{l}$, and $a_{sj}=\vec{q}_{1j} \cdot \vec{q}_{2s}$, $b_{sl}=\vec{p}_{l} \cdot \vec{q}_{2s}$ then expression \ref{Ex:combinePrac} is 
\begin{align} 
&\vec{w_i} - \sum_{j}^{r_1}(\vec{q}_{1j} \cdot \vec{w}_i)\vec{q}_{1j} - \sum_{s}^{r_2}(\sum_{j}^{r_1}a_{sj}\vec{q}_{1j}+
\sum_{l}^{r_m - r_1}b_{sl}\vec{p}_{l} \cdot \vec{w}_i)\vec{q}_{2s} +
\sum_{s}^{r_2}((\sum_{j}^{r_1}(\vec{q}_{1j} \cdot \vec{w}_i)\vec{q}_{1j})\cdot \vec{q}_{2s})\vec{q}_{2s}
\\
&=\vec{w_i} - \sum_{j}^{r_1}(\vec{q}_{1j} \cdot \vec{w}_i)\vec{q}_{1j} - \sum_{s}^{r_2}((\sum_{j}^{r_1}a_{sj}\vec{q}_{1j}+
\sum_{l}^{r_m - r_1}b_{sl}\vec{p}_{l} )\cdot \vec{w}_i)\vec{q}_{2s} +
\sum_{s}^{r_2}((\sum_{j}^{r_1}(\vec{q}_{1j} \cdot \vec{w}_i)\vec{q}_{1j})\cdot \vec{q}_{2s})\vec{q}_{2s} \\
&=... - \sum_{s}^{r_2}((\sum_{j}^{r_1}a_{sj}\vec{q}_{1j}\cdot \vec{w}_i) -
((\sum_{j}^{r_1}(\vec{q}_{1j} \cdot \vec{w}_i)\vec{q}_{1j})\cdot \vec{q}_{2s}))\vec{q}_{2s}+
\sum_{s}^{r_2}\sum_{l}^{r_m - r_1}(b_{sl}\vec{p}_{l} \cdot \vec{w}_i)\vec{q}_{2s}
\end{align} 
When substituting $a_{sj}$ and $b_{sl}$, the intermediate terms are eliminated, which is 
\begin{align}
&\vec{w_i} - \sum_{j}^{r_1}(\vec{q}_{1j} \cdot \vec{w}_i)\vec{q}_{1j} + \sum_{s}^{r_2}\sum_{l}^{r_m - r_1}(\vec{p}_{l} \cdot \vec{q}_{2s} \vec{p}_{l} \cdot \vec{w}_i)\vec{q}_{2s}\\
&=\vec{w_i} - \sum_{j}^{r_1}(\vec{q}_{1j} \cdot \vec{w}_i)\vec{q}_{1j} + \sum_{l}^{r_m - r_1}\sum_{s}^{r_2}(\vec{p}_{l} \cdot \vec{q}_{2s} \vec{q}_{2s} )\vec{p}_{l} \cdot \vec{w}_i
\label{Ex:combinePrac_final}
\end{align} 

The term $\sum_{s}^{r_2}(\vec{p}_{l} \cdot \vec{q}_{2s} \vec{q}_{2s} )$ represents the total component of $\vec{p}_{l}$ in the column space of $Q_2$. As $\vec{p}_{l}$  is orthogonal with $\{\vec{q}_{11}, \vec{q}_{12} ... \vec{q}_{1r_1}\}$, we have $\sum_{s}^{r_2}(\vec{p}_{l} \cdot \vec{q}_{2s} \vec{q}_{2s} ) = \vec{p}_{l}$. Hence, it is proven that Expression \ref{Ex:combineTheory} and Expression \ref{Ex:combinePrac_final} are equal.  

% \begin{align} 
% \vec{w_i} - 
% \sum_{j}^{r_1}(\vec{q}_{1j} \cdot \vec{w}_i)\vec{q}_{1j} -\\
% \sum_{s}^{r_2}((\sum_{j}^{r_1}(\vec{q}_{1j} \cdot \vec{q}_{2s})\vec{q}_{1j}+
% \sum_{l}^{r_m - r_1}(\vec{p}_{l} \cdot \vec{q}_{2s})\vec{p}_{l}) \cdot \vec{w}_i)(\sum_{j}^{r_1}(\vec{q}_{1j} \cdot \vec{q}_{2s})\vec{q}_{1j}+
% \sum_{l}^{r_m - r_1}(\vec{p}_{l} \cdot \vec{q}_{2s})\vec{p}_{l}) +\\
% \sum_{s}^{r_2}(\sum_{j}^{r_1}(\vec{q}_{1j} \cdot \vec{w}_i)\vec{q}_{1j})\vec{q}_{2s}
% \end{align} 

\section{Comparisons of the Generation Performance for Different Subset Parameters}
\label{sec:appendixC}
\begin{figure}
% \vskip -0.1in
  \centering
\centerline{\includegraphics[width=1.0\columnwidth]{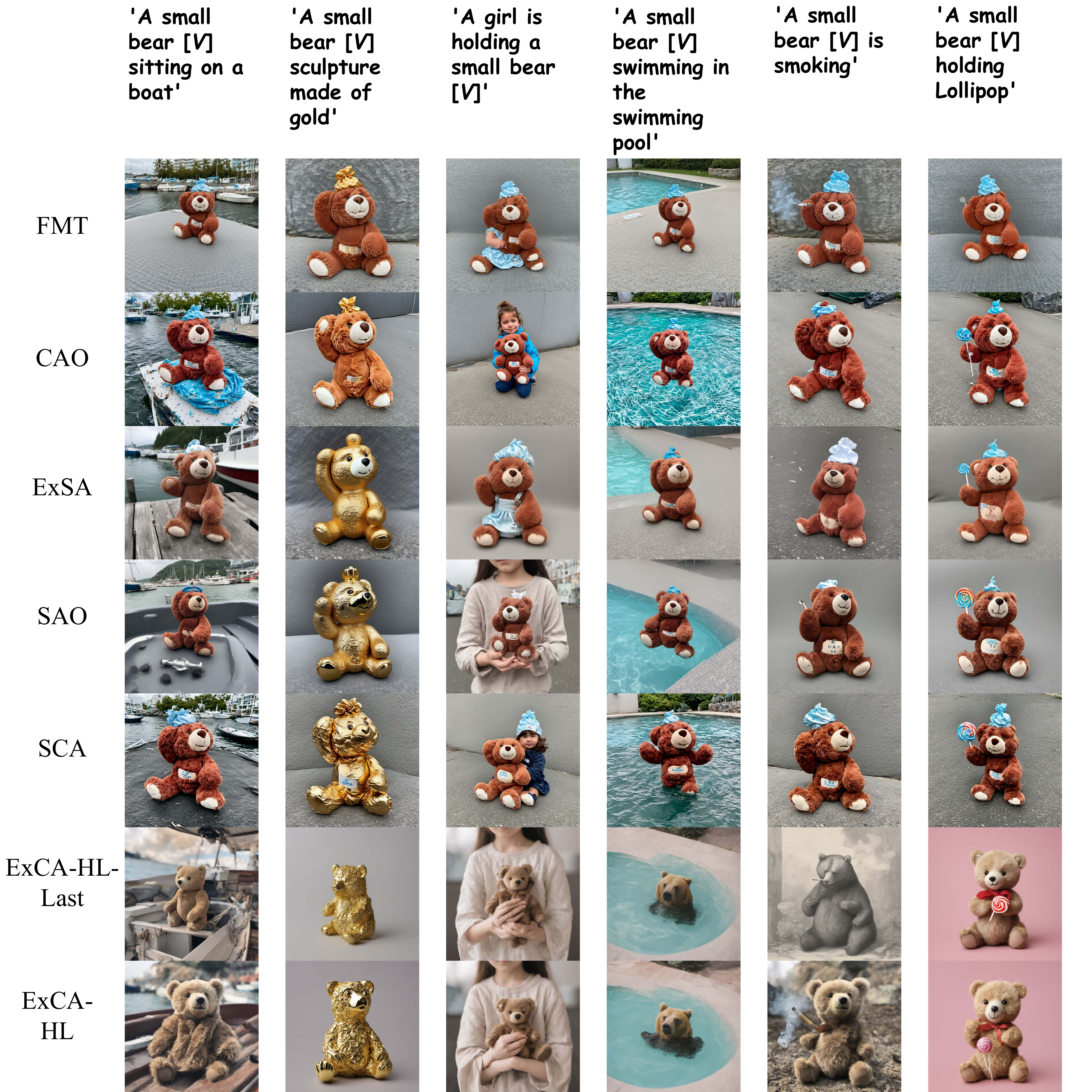}}
  \caption{PaRa Single subject generation on different parameter subsets}
\label{more_subset_singleimagegenerate}
\end{figure}

ExCA, as a parameter subset,  is choosed in our experiments. Here, we provide additional examples generated using other subsets of parameters. We experimented with eight different parameter subsets. 
\begin{itemize}
    \item \textbf{Exclude Cross-Attention (ExCA)}
        \begin{itemize}
            \item This method trains all layers except cross-attention and time embedding layers.
        \end{itemize}
    \item \textbf{Exclude Self-Attention (ExSA)}
        \begin{itemize}
            \item This method trains all layers except self-attention layers.
        \end{itemize}
    \item \textbf{Self-Attention Only (SAO)}
        \begin{itemize}
            \item This method trains only the self-attention layers.
        \end{itemize}
    \item \textbf{Cross-Attention Only (CAO)}
        \begin{itemize}
            \item This method trains only the cross-attention layers.
        \end{itemize}
    \item \textbf{Full Model Training (FMT)}
        \begin{itemize}
            \item This method trains all layers of the model.
        \end{itemize}
    \item \textbf{Strict Cross-Attention (SCA)}
        \begin{itemize}
            \item This method trains only the queries and keys within the cross-attention mechanisms.
        \end{itemize}
    \item \textbf{Exclude Cross-Attention High-Level (ExCA-HL)}
        \begin{itemize}
            \item This method trains all layers except cross-attention layers, with an emphasis on high-level feature representations.
        \end{itemize}
    \item \textbf{Exclude Cross-Attention High-Level Last (ExCA-HL-Last)}
        \begin{itemize}
            \item This method trains all layers except cross-attention layers, focusing on the final stages of the high-level feature space.
        \end{itemize}
\end{itemize}
As shown in Table \ref{modelsize_appendix}, regardless of the subset used, the parameter count of PaRa is significantly lower than that of LoRA. In Fig. \ref{more_subset_singleimagegenerate}, we use the example of a bear plushie, with $r=16$, to illustrate the effects of different parameter subsets. It can be observed that models with larger parameter counts, such as FMT, CAO, and ExSA, tend to align better with the training images. Conversely, models with smaller parameter counts may not align well with the target subject but match the text more closely. The model with the largest parameter count, FMT, even produced a 'hybrid' result in the multi-subject example "A girl is holding a small bear [V]". Models like SAO and SCA strike a better balance. EXCA, being a well-performing subset, has numerous examples listed in other sections and will not be repeated here.

\begin{table}[h]
% // a headshot of a person $<$Male$>$ $<$Skin tone 5$>$ $<$Age 60 69$>$}
% \vskip -0.2in
\begin{center}
\begin{small}
\begin{sc}
\resizebox{1.0\linewidth}{!}{
\begin{tabular}{@{}ccccccccc@{}} 
\toprule
& \multicolumn{2}{c}{$r$=2}  &&  \multicolumn{2}{c}{$r$=16} &&  \multicolumn{2}{c}{$r$=128} \\
\midrule
Subset & PaRa & LoRA && PaRa & LoRA && PaRa & LoRA \\
\midrule
Exclude Cross-Attention (ExCA) & 1.8 MB &4.8 MB && 13 MB &33 MB && 87 MB &190 MB\\
Exclude Self-Attention (ExSA)  &1.9 MB &4.8 MB &&13 MB &33 MB &&87 MB &190 MB\\
Self-Attention Only (SAO) & 1.6M &3.1M &&11M &21M &&82M & 163M\\
Cross-Attention Only (CAO)    & 1.6M & 3.5M && 11M & 25M && 82M & 193M\\
Full Model Training (FMT)    &3.4M & 8.2M && 23M & 58M && 169M & 382M\\
Strict Cross-Attention (SCA)   & 1.2M & 2.8M && 7.9M & 20M && 62M & 152M\\
Exclude Cross-Attention High-Level (ExCA-HL)&276K & 700K && 1.9M & 5.0M && 14M & 30M\\
Exclude Cross-Attention High-Level Last (ExCA-HL-Last)& 6.9K & 53K && 42K & 403K && 82K & 803K\\
\bottomrule
\end{tabular}
}
\end{sc}
\end{small}
\end{center}
\caption{Fine-tuning subsets of parameters in UNet, comparing them with LoRA at various ranks, along with their corresponding model sizes.}
\label{modelsize_appendix}
% \vskip -0.4in
\end{table}

\section{Comparison of Different Rank Boundaries $\gamma$}
\label{sec:appendixD}
\begin{figure}
% \vskip -0.1in
  \centering
\centerline{\includegraphics[width=1.0\columnwidth]{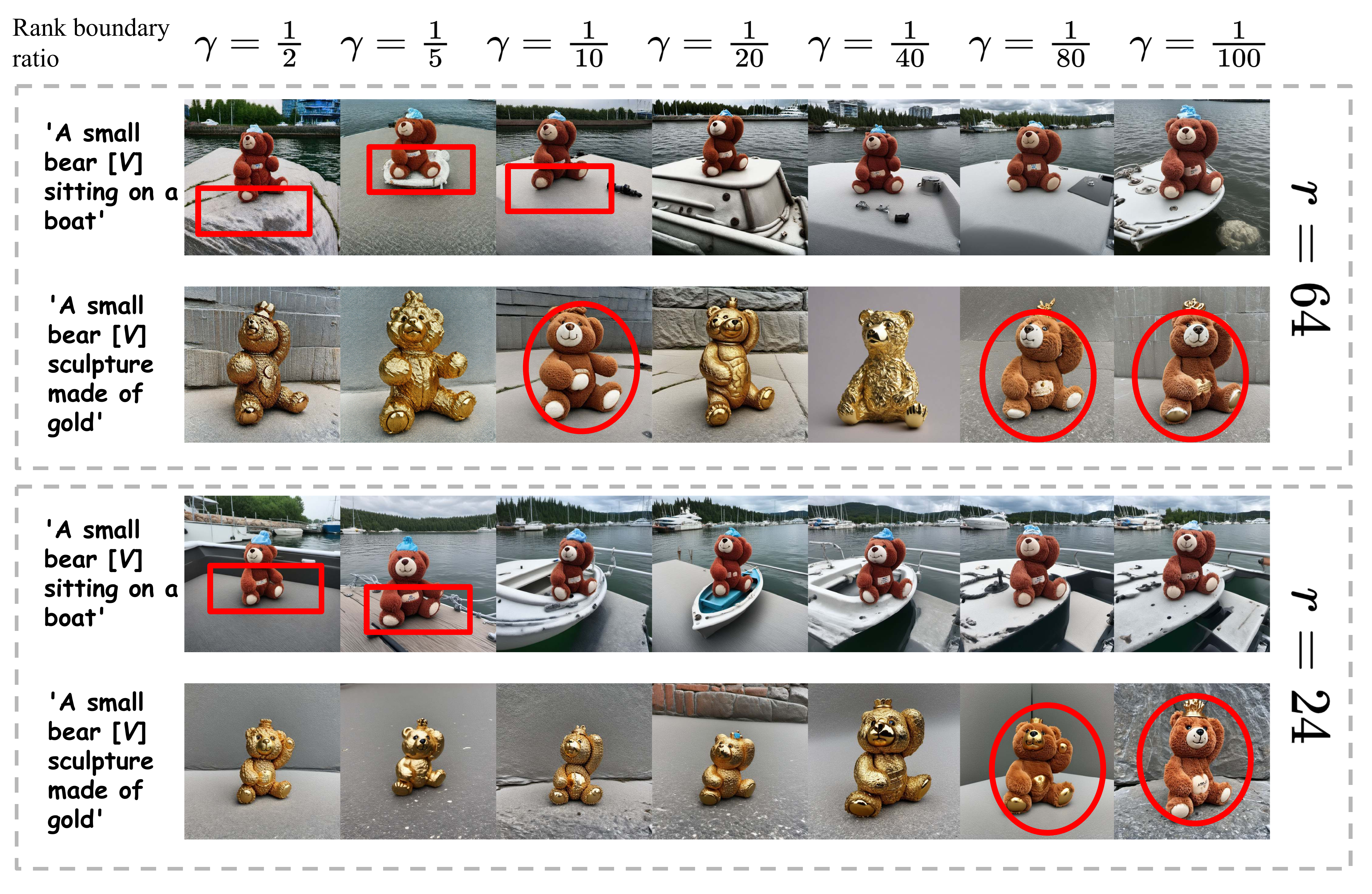}}
  \caption{Comparison of Different Rank Boundaries $\gamma$ for single subject generation. Red boxes indicate errors in generating ``boat'', and red circles indicate errors in generating ``gold''.}
\label{fig:more_rankratio}
\end{figure}

In Section \ref{sec:CombiningPaRa}, we discussed that setting the $r$ too high can lead to model instability. To mitigate this issue, it is necessary to establish a rank boundary $\gamma$. We empirically chose $\gamma=\frac{1}{40}$ across various experimental scenarios. The Fig. \ref{fig:more_rankratio} illustrates the results generated with different $\gamma$ values. Red markers indicate errors in the generated images. The figure demonstrates that when $r$ is small, the setting of $\gamma$ has minimal impact. However, as $r$ increases, it becomes crucial to select an appropriate $\gamma$, with values around $\frac{1}{20}$ to $\frac{1}{40}$ yielding the best performance.

\section{Different $r$ Values in Image Editing and Model Combination}
\label{sec:appendixE}
\begin{figure}
% \vskip -0.1in
  \centering
\centerline{\includegraphics[width=1.0\columnwidth]{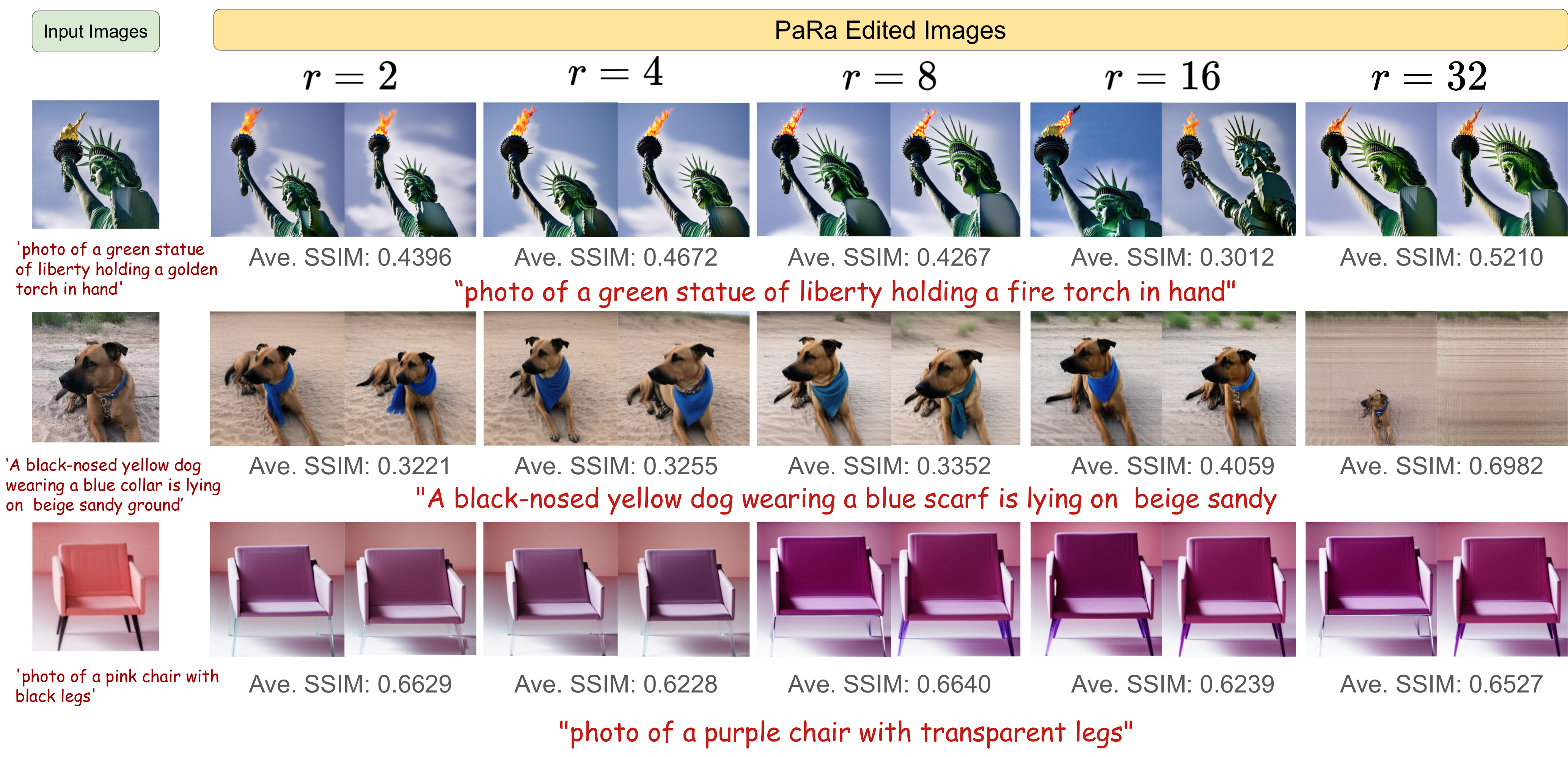}}
  \caption{Comparison of Different Rank Image Editing.$r=2$ to $r=8$ show minimal differences. When $r$ is larger, such as $r=16$ to $32$, the generated results for more challenging images tend to degrade significantly.}
\label{fig:Rank_ImageEdit_compare_appendix}
\end{figure}

In Figure \ref{fig:Rank_ImageEdit_compare_appendix}, we compare the performance of PaRa in one-shot learning for image editing across different rank values. We also calculate the average SSIM for each prompt generated with different ranks. It can be observed that the results for $r=2$ to $r=8$ show minimal differences. However, when $r$ is larger, such as $r=16$ to $32$, the generated results for more challenging images tend to degrade significantly.

\begin{figure}
% \vskip -0.1in
  \centering
\centerline{\includegraphics[width=1.0\columnwidth]{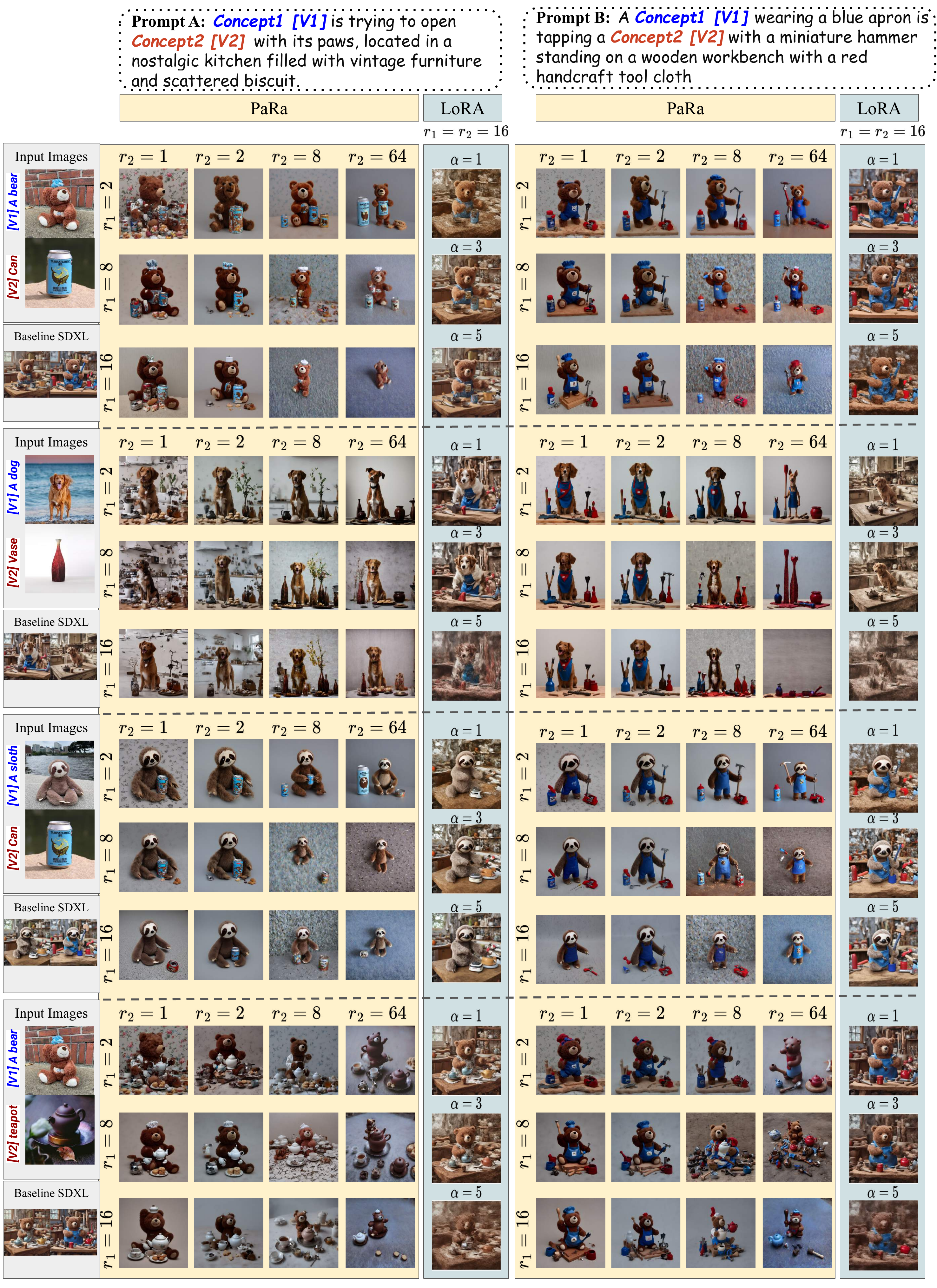}}
  \caption{Multi-Subject Generation. In the PaRa comparison, we analyzed results for Concept 1 with reduced ranks $r_1$ set at 2, 8, 16; and for Concept 2, the reduced ranks $r_2$ were 1, 2, 8, 64. LoRA, based on experimental optimization, used the best rank of 16 for both concepts and compared different scales with alpha values of 1, 3, 5.}
\label{fig:multigenerate_appendix}
\end{figure}

In Figure \ref{fig:multigenerate_appendix}, we compare the performance of PaRa with different ranks in model combination. The results indicate that ranks between 2 and 8 are more suitable. When the rank $r$ is too large, the interaction between the two PaRa models becomes significant, potentially causing one subject to be ignored in the generated results. However, if there is a primary and secondary subject, the rank $r$ of the PaRa corresponding to the primary subject can be set higher.

\section{More Generation Results}
\label{sec:appendixF}
\begin{figure}
% \vskip -0.1in
  \centering
\centerline{\includegraphics[width=1.0\columnwidth]{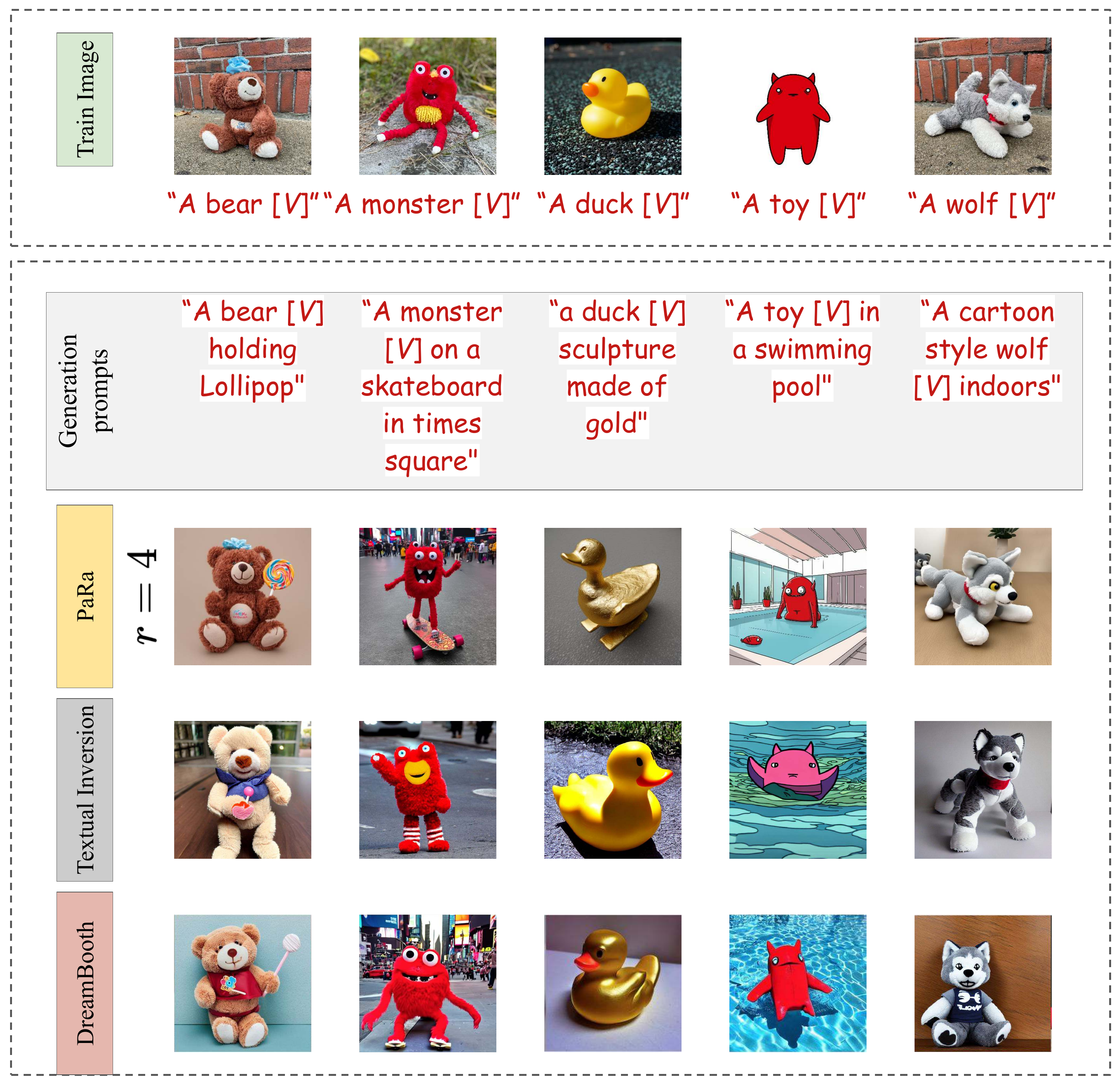}}
  \caption{Comparison of Single-Subject Generation with DreamBooth and Textual Inversion}
\label{singleimagegenerate_dreambooth_and_textual_inversion}
\end{figure}

In Figure \ref{singleimagegenerate_dreambooth_and_textual_inversion}, we compare the performance of PaRa in single-subject generation with DreamBooth and Textual Inversion. The results clearly demonstrate the significant advantages of PaRa.

\begin{figure}[h]

\centering            \includegraphics[width=1.0\linewidth]{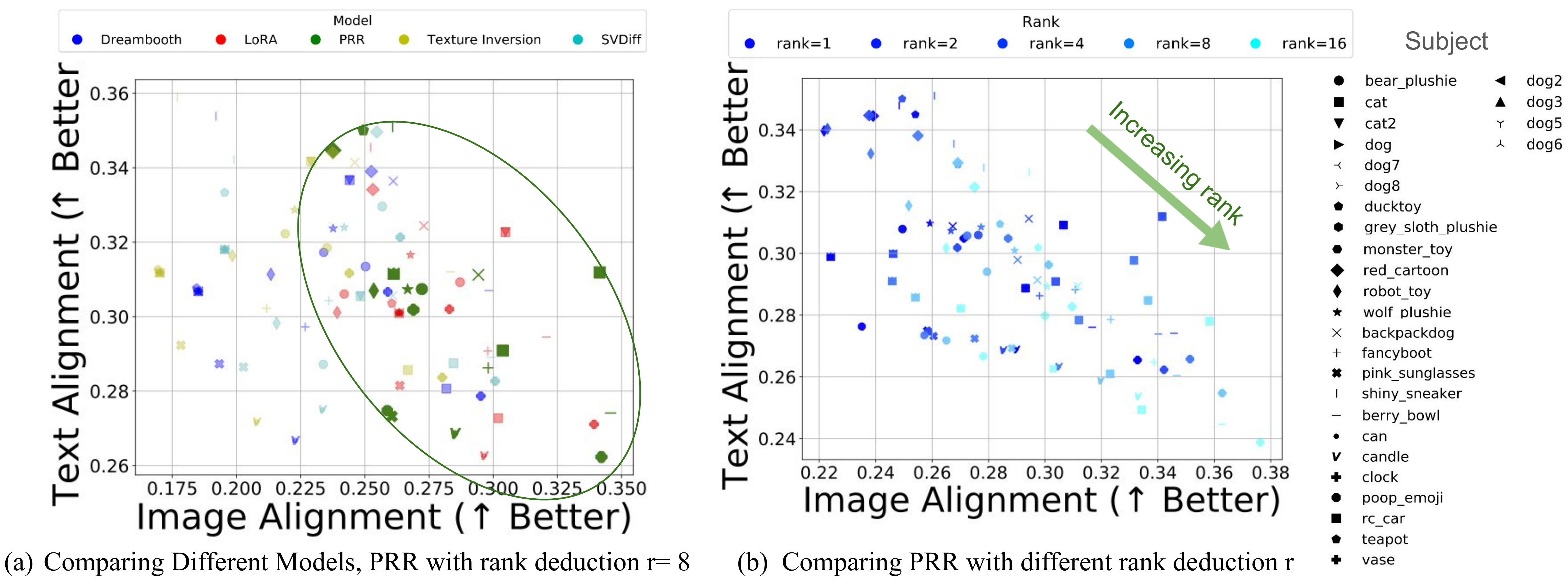}

\caption{Text and image alignment for single subject generation. The generated image is denoted as $\tilde{\mathbf{x}}$, the prompt is denoted as $c$, the train image is denoted as $\mathbf{x}$. The text alignment is measured by the CLIP score $cos(\tilde{\mathbf{x}}, c)$, and the image alignment is defined as $1-\mathcal{L}_{\text {LPIPS }}\left(\tilde{\mathbf{x}}, \mathbf{x}\right)$. In figure (a), our model PaRa is positioned in the lower left corner in the green circle. In figure (b), the green arrow indicates the approximate direction of rank increase.}
% \vskip -0.1in
\label{fig:text_image_alignment_appendix}
% \vskip -0.1in
\end{figure}

Figure \ref{fig:text_image_alignment_appendix} serves as a supplementary figure to Figure \ref{fig:text_image_alignment} in the main text, providing benchmarks of the text and image alignment for each subject.

\begin{figure}
% \vskip -0.1in
  \centering
\centerline{\includegraphics[width=1.0\columnwidth]{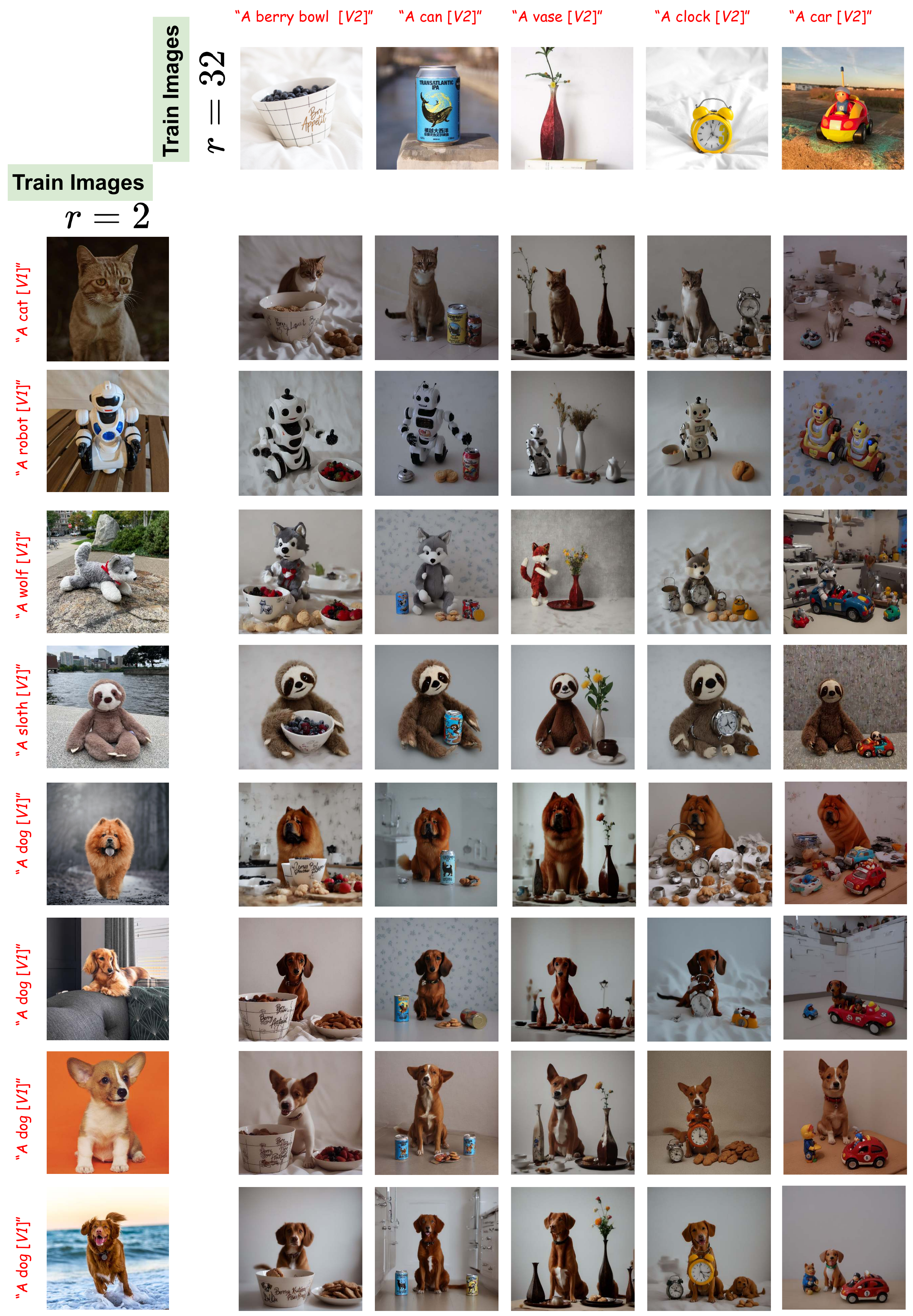}}
  \caption{More Multi-Subject Generation Examples: The left column shows the training images of the primary subject, and the top row shows the training images of the secondary subject, each represented by a single image but actually based on five images for few-shot training. The prompts used here are ``[\textit{V1}] in the kitchen with a [\textit{V2}] next to it.''
  % \textcolor{red}{“[\textit{V1}] in the kitchen with a [\textit{V2}] next to it.”}
  }
\label{fig:MorePaRaCombineDemo}
\end{figure}

In Figure \ref{fig:MorePaRaCombineDemo}, we present more examples of PaRa combinations. In these examples, the primary subject rank is 2, and the secondary subject rank is 32, which we have found through experience to be a relatively suitable pairing.

\section{Image Editing Algebraic Discussion}
\label{sec:appendixG}
The stability of PaRa outputs is reflected in that different Gaussian noises tend to yield the same result, represented as:

\begin{align} \label{eq:kernel}
h = Wx = W(x+\Delta x) 
\end{align} 
the equation is true when $W\Delta x=\mathbf{0} $, which is denoted by $\Delta x \in  kernel(W)$.

According to the rank-nullity theorem, for the linear transformation $W:X\to H$, ${rank(W)+nullity(W)=\dim X}$. (The nullity of $W$ is the dimension of $W$ kernel).) In PaRa, our reduced rank $r=rank(W_0) - rank(W_{reduce})=nullity(W_{reduce})-nullity(W_0)$, we have $rank(W_0) > rank(W_{reduce})$ which implies  $nullity(W_{reduce}) > nullity(W_0)$, more of $\Delta x$ in PaRa will not produce different outputs.
 This reduced rank $r$ demonstrates the problem \citep{meng2021sdedit} of the trade-off between faithful reconstruction and editability in image editing, As $r$ increases, the modifiable features decrease, making the reconstruction more faithful. As $r$ decreases, the modifiable features increase, bringing the diversity of the generated images closer to that of the underlying pre-trained generation model, and improving the editability. When a large $r$ is selected for training the PaRa on a single image, the model generates images that closely resemble the training image, even when using various text prompts. This enables direct modification of the text prompt to facilitate image editing on the single train image. 

\section{Survey: User Assessment of Single-Subject and Multi-Subject Image Generation}
\label{sec:appendixH}
% \textbf{Human Reflection on PaRa Performance}  
\begin{table}[t]
% // a headshot of a person $<$Male$>$ $<$Skin tone 5$>$ $<$Age 60 69$>$}
% \vskip -0.2in
\begin{center}
\begin{small}
\begin{sc}
\resizebox{1.0\linewidth}{!}{
\begin{tabular}{@{}cccccccccc@{}} 
\toprule
 & bear\_plushie & red\_toy & ducktoy & red\_cartoon & wolf\_plushie & bear\_plushie\&tin & dog\&vase & cat\&teapot &  grey\_sloth\_plushie\&tin\\
% & \multicolumn{3}{c}{cat}  &&  \multicolumn{3}{c}{dog} &&  \multicolumn{3}{c}{ducktoy} \\
\midrule
Human Preference PaRa& 96.17\%&97.13\%&91.18\%&91.90\%&96.62\%&82.08\%&89.94\%&78.45\%&88.76\%\\
Human Preference LoRA& 3.83\%&2.87\%&8.82\%&8.09\%&3.38\%&7.54\%&3.91\%&15.47\%&7.30\%\\
\bottomrule
\end{tabular}
}
\end{sc}
\end{small}
\end{center}
\caption{Human evaluation results comparing PaRa and LoRA. The first five columns compare single-subject generation, while the last four columns compare multi-subject combination. Each of the last four columns does not sum to $100\%$ because there is a ``Neither'' option in the survey indicating dissatisfaction with both results.}
\label{tab:Survey}
% \vskip -0.4in
\end{table}

To evaluate the quality of the generated images, we conducted a survey where 209 participants provided their feedback. In order to streamline and ensure effectiveness, we designed a survey form consisting of nine questions. These questions allowed users to compare the image generation performance of PaRa and LoRA, covering both single-subject and multi-subject generation, as well as comparisons across different ranks. 
% The overall results are summarized in Table \ref{tab:Survey}. 
% The link of the survey form and the statistical results are provided in the Appendix \ref{sec:appendixH}.

The survey form is shown in this link: 

\url{https://docs.google.com/forms/d/e/1FAIpQLSc7mDMagFkotVwYiYjftV3FY_WSoYDtcpX6_3VqVlp9SmREbA/viewform?usp=sf_link}.

% \url{https://wj.qq.com/s2/14673105/0a28/}.

If you click on the survey link, you will find that participants are blind to which model corresponds to each image, which helps reduce subjective bias. The survey may have regional bias because 181 participants completed the Google Form, while 28 participants completed the Tencent Form. The survey consists of 9 questions. The first five questions are actually about PaRa and LoRA's Single-Subject generation. Options (a)-(e) correspond to PaRa ranks 1, 2, 4, 8, and 16, respectively. Options (f) and (g) correspond to LoRA rank 8 with scales 1.0 and 2.2, respectively. The last four questions are about multi-subject generation, where options (a) and (b) correspond to the results of PaRa and LoRA, respectively.

The overall results are summarized in Table \ref{tab:Survey}.

%% file: main.bbl
\begin{thebibliography}{47}
\providecommand{\natexlab}[1]{#1}
\providecommand{\url}[1]{\texttt{#1}}
\expandafter\ifx\csname urlstyle\endcsname\relax
  \providecommand{\doi}[1]{doi: #1}\else
  \providecommand{\doi}{doi: \begingroup \urlstyle{rm}\Url}\fi

\bibitem[dia(2024)]{diablo}
diablo-sliders.
\newblock \url{https://civitai.com/models/228180/diablo-sliders-ntcaixyz},
  2024.

\bibitem[fig(2024)]{figurine}
Figurine-sliders.
\newblock \url{https://civitai.com/models/231983/figurine-sliders-ntcaixyz},
  2024.

\bibitem[mec(2024)]{mecha}
mecha-slider.
\newblock \url{https://civitai.com/models/317484/mecha-slider-sdxl}, 2024.

\bibitem[Andrews and Patterson(1976)]{andrews1976singular}
H~Andrews and CLIII Patterson.
\newblock Singular value decomposition (svd) image coding.
\newblock \emph{IEEE transactions on Communications}, 24\penalty0 (4):\penalty0
  425--432, 1976.

\bibitem[Chang et~al.(2023)Chang, Zhang, Barber, Maschinot, Lezama, Jiang,
  Yang, Murphy, Freeman, Rubinstein, et~al.]{chang2023muse}
Huiwen Chang, Han Zhang, Jarred Barber, AJ~Maschinot, Jose Lezama, Lu~Jiang,
  Ming-Hsuan Yang, Kevin Murphy, William~T Freeman, Michael Rubinstein, et~al.
\newblock Muse: Text-to-image generation via masked generative transformers.
\newblock \emph{arXiv preprint arXiv:2301.00704}, 2023.

\bibitem[Chen et~al.(2024)Chen, Hu, Li, Ruiz, Jia, Chang, and
  Cohen]{chen2024subject}
Wenhu Chen, Hexiang Hu, Yandong Li, Nataniel Ruiz, Xuhui Jia, Ming-Wei Chang,
  and William~W Cohen.
\newblock Subject-driven text-to-image generation via apprenticeship learning.
\newblock \emph{Advances in Neural Information Processing Systems}, 36, 2024.

\bibitem[cloneofsimo(2022)]{lora_git}
cloneofsimo.
\newblock Low-rank adaptation for fast text-to-image diffusion fine-tuning.
\newblock \url{https://github.com/cloneofsimo/lora}, 2022.

\bibitem[Feng et~al.(2022)Feng, Zheng, Huang, Zhao, Jordan, and
  Zha]{feng2022rank}
Ruili Feng, Kecheng Zheng, Yukun Huang, Deli Zhao, Michael Jordan, and
  Zheng-Jun Zha.
\newblock Rank diminishing in deep neural networks.
\newblock \emph{Advances in Neural Information Processing Systems},
  35:\penalty0 33054--33065, 2022.

\bibitem[Gal et~al.(2022)Gal, Alaluf, Atzmon, Patashnik, Bermano, Chechik, and
  Cohen-Or]{textinversion}
Rinon Gal, Yuval Alaluf, Yuval Atzmon, Or~Patashnik, Amit~H Bermano, Gal
  Chechik, and Daniel Cohen-Or.
\newblock An image is worth one word: Personalizing text-to-image generation
  using textual inversion.
\newblock \emph{arXiv preprint arXiv:2208.01618}, 2022.

\bibitem[Gal et~al.(2023)Gal, Arar, Atzmon, Bermano, Chechik, and
  Cohen{-}Or]{gal2023designing}
Rinon Gal, Moab Arar, Yuval Atzmon, Amit~H. Bermano, Gal Chechik, and Daniel
  Cohen{-}Or.
\newblock Encoder-based domain tuning for fast personalization of text-to-image
  models.
\newblock \emph{{ACM} Trans. Graph.}, 2023.

\bibitem[Gandikota et~al.(2023)Gandikota, Materzynska, Zhou, Torralba, and
  Bau]{gandikota2023concept}
Rohit Gandikota, Joanna Materzynska, Tingrui Zhou, Antonio Torralba, and David
  Bau.
\newblock Concept sliders: Lora adaptors for precise control in diffusion
  models.
\newblock \emph{arXiv preprint arXiv:2311.12092}, 2023.

\bibitem[Gu et~al.(2022)Gu, Chen, Bao, Wen, Zhang, Chen, Yuan, and
  Guo]{gu2022vector}
Shuyang Gu, Dong Chen, Jianmin Bao, Fang Wen, Bo~Zhang, Dongdong Chen, Lu~Yuan,
  and Baining Guo.
\newblock Vector quantized diffusion model for text-to-image synthesis.
\newblock In \emph{Proceedings of the IEEE/CVF Conference on Computer Vision
  and Pattern Recognition}, pages 10696--10706, 2022.

\bibitem[Gu et~al.(2024)Gu, Wang, Wu, Shi, Chen, Fan, Xiao, Zhao, Chang, Wu,
  et~al.]{gu2024mix}
Yuchao Gu, Xintao Wang, Jay~Zhangjie Wu, Yujun Shi, Yunpeng Chen, Zihan Fan,
  Wuyou Xiao, Rui Zhao, Shuning Chang, Weijia Wu, et~al.
\newblock Mix-of-show: Decentralized low-rank adaptation for multi-concept
  customization of diffusion models.
\newblock \emph{Advances in Neural Information Processing Systems}, 36, 2024.

\bibitem[Guo et~al.(2016)Guo, Yao, and Chen]{guo2016dynamic}
Yiwen Guo, Anbang Yao, and Yurong Chen.
\newblock Dynamic network surgery for efficient dnns.
\newblock \emph{Advances in neural information processing systems}, 29, 2016.

\bibitem[Han et~al.(2023)Han, Li, Zhang, Milanfar, Metaxas, and
  Yang]{han2023svdiff}
Ligong Han, Yinxiao Li, Han Zhang, Peyman Milanfar, Dimitris Metaxas, and Feng
  Yang.
\newblock Svdiff: Compact parameter space for diffusion fine-tuning.
\newblock In \emph{Proceedings of the IEEE/CVF International Conference on
  Computer Vision}, pages 7323--7334, 2023.

\bibitem[Han et~al.(2015)Han, Pool, Tran, and Dally]{han2015learning}
Song Han, Jeff Pool, John Tran, and William Dally.
\newblock Learning both weights and connections for efficient neural network.
\newblock \emph{Advances in neural information processing systems}, 28, 2015.

\bibitem[Ho et~al.(2020)Ho, Jain, and Abbeel]{ho2020denoising}
Jonathan Ho, Ajay Jain, and Pieter Abbeel.
\newblock Denoising diffusion probabilistic models.
\newblock \emph{Advances in neural information processing systems},
  33:\penalty0 6840--6851, 2020.

\bibitem[Hu et~al.(2021)Hu, Shen, Wallis, Allen-Zhu, Li, Wang, Wang, and
  Chen]{hu2021lora}
Edward~J Hu, Yelong Shen, Phillip Wallis, Zeyuan Allen-Zhu, Yuanzhi Li, Shean
  Wang, Lu~Wang, and Weizhu Chen.
\newblock Lora: Low-rank adaptation of large language models.
\newblock \emph{arXiv preprint arXiv:2106.09685}, 2021.

\bibitem[Idelbayev and Carreira{-}Perpi{\~{n}}{\'{a}}n(2020)]{IdelbayevC20}
Yerlan Idelbayev and Miguel~{\'{A}}. Carreira{-}Perpi{\~{n}}{\'{a}}n.
\newblock Low-rank compression of neural nets: Learning the rank of each layer.
\newblock In \emph{CVPR}, pages 8046--8056, 2020.

\bibitem[Jia et~al.(2023)Jia, Zhao, Chan, Li, Zhang, Gong, Hou, Wang, and
  Su]{jia2023taming}
Xuhui Jia, Yang Zhao, Kelvin~CK Chan, Yandong Li, Han Zhang, Boqing Gong,
  Tingbo Hou, Huisheng Wang, and Yu-Chuan Su.
\newblock Taming encoder for zero fine-tuning image customization with
  text-to-image diffusion models.
\newblock \emph{arXiv preprint arXiv:2304.02642}, 2023.

\bibitem[Kumari et~al.(2023)Kumari, Zhang, Zhang, Shechtman, and
  Zhu]{kumari2023multi}
Nupur Kumari, Bingliang Zhang, Richard Zhang, Eli Shechtman, and Jun-Yan Zhu.
\newblock Multi-concept customization of text-to-image diffusion.
\newblock In \emph{Proceedings of the IEEE/CVF Conference on Computer Vision
  and Pattern Recognition}, pages 1931--1941, 2023.

\bibitem[Liu et~al.(2015)Liu, Wang, Foroosh, Tappen, and Pensky]{liu2015sparse}
Baoyuan Liu, Min Wang, Hassan Foroosh, Marshall Tappen, and Marianna Pensky.
\newblock Sparse convolutional neural networks.
\newblock In \emph{Proceedings of the IEEE conference on computer vision and
  pattern recognition}, pages 806--814, 2015.

\bibitem[Meng et~al.(2021)Meng, Song, Song, Wu, Zhu, and Ermon]{meng2021sdedit}
Chenlin Meng, Yang Song, Jiaming Song, Jiajun Wu, Jun-Yan Zhu, and Stefano
  Ermon.
\newblock Sdedit: Image synthesis and editing with stochastic differential
  equations.
\newblock \emph{arXiv preprint arXiv:2108.01073}, 2021.

\bibitem[{MidJourney}()]{midjourney}
{MidJourney}.
\newblock Midjourney.
\newblock \url{https://www.midjourney.com}.

\bibitem[Mokady et~al.(2023)Mokady, Hertz, Aberman, Pritch, and
  Cohen-Or]{mokady2023null}
Ron Mokady, Amir Hertz, Kfir Aberman, Yael Pritch, and Daniel Cohen-Or.
\newblock Null-text inversion for editing real images using guided diffusion
  models.
\newblock In \emph{Proceedings of the IEEE/CVF Conference on Computer Vision
  and Pattern Recognition}, pages 6038--6047, 2023.

\bibitem[Mou et~al.(2024)Mou, Wang, Xie, Wu, Zhang, Qi, and Shan]{mou2024t2i}
Chong Mou, Xintao Wang, Liangbin Xie, Yanze Wu, Jian Zhang, Zhongang Qi, and
  Ying Shan.
\newblock T2i-adapter: Learning adapters to dig out more controllable ability
  for text-to-image diffusion models.
\newblock In \emph{Proceedings of the AAAI Conference on Artificial
  Intelligence}, volume~38, pages 4296--4304, 2024.

\bibitem[Nichol and Dhariwal(2021)]{nichol2021improved}
Alexander~Quinn Nichol and Prafulla Dhariwal.
\newblock Improved denoising diffusion probabilistic models.
\newblock In \emph{International conference on machine learning}, pages
  8162--8171. PMLR, 2021.

\bibitem[OpenAI(2023)]{dalle3}
OpenAI.
\newblock Dall·e 3: Language models for image generation.
\newblock \url{https://cdn.openai.com/papers/dall-e-3.pdf}, 2023.

\bibitem[Podell et~al.(2023)Podell, English, Lacey, Blattmann, Dockhorn,
  M{\"u}ller, Penna, and Rombach]{sdxl}
Dustin Podell, Zion English, Kyle Lacey, Andreas Blattmann, Tim Dockhorn, Jonas
  M{\"u}ller, Joe Penna, and Robin Rombach.
\newblock Sdxl: Improving latent diffusion models for high-resolution image
  synthesis.
\newblock \emph{arXiv preprint arXiv:2307.01952}, 2023.

\bibitem[Radford et~al.(2021)Radford, Kim, Hallacy, Ramesh, Goh, Agarwal,
  Sastry, Askell, Mishkin, Clark, et~al.]{radford2021learning}
Alec Radford, Jong~Wook Kim, Chris Hallacy, Aditya Ramesh, Gabriel Goh,
  Sandhini Agarwal, Girish Sastry, Amanda Askell, Pamela Mishkin, Jack Clark,
  et~al.
\newblock Learning transferable visual models from natural language
  supervision.
\newblock In \emph{International conference on machine learning}, pages
  8748--8763. PMLR, 2021.

\bibitem[Ramesh et~al.(2022)Ramesh, Dhariwal, Nichol, Chu, and
  Chen]{ramesh2022hierarchical}
Aditya Ramesh, Prafulla Dhariwal, Alex Nichol, Casey Chu, and Mark Chen.
\newblock Hierarchical text-conditional image generation with clip latents.
\newblock \emph{arXiv preprint arXiv:2204.06125}, 2022.

\bibitem[Robb et~al.(2020)Robb, Chu, Kumar, and Huang]{robb2020few}
Esther Robb, Wen-Sheng Chu, Abhishek Kumar, and Jia-Bin Huang.
\newblock Few-shot adaptation of generative adversarial networks.
\newblock \emph{arXiv preprint arXiv:2010.11943}, 2020.

\bibitem[Rombach et~al.(2022)Rombach, Blattmann, Lorenz, Esser, and Ommer]{sd}
Robin Rombach, Andreas Blattmann, Dominik Lorenz, Patrick Esser, and Bj{\"o}rn
  Ommer.
\newblock High-resolution image synthesis with latent diffusion models.
\newblock In \emph{Proceedings of the IEEE/CVF conference on computer vision
  and pattern recognition}, pages 10684--10695, 2022.

\bibitem[Ruiz et~al.(2023)Ruiz, Li, Jampani, Pritch, Rubinstein, and
  Aberman]{dreambooth}
Nataniel Ruiz, Yuanzhen Li, Varun Jampani, Yael Pritch, Michael Rubinstein, and
  Kfir Aberman.
\newblock Dreambooth: Fine tuning text-to-image diffusion models for
  subject-driven generation.
\newblock In \emph{Proceedings of the IEEE/CVF Conference on Computer Vision
  and Pattern Recognition}, pages 22500--22510, 2023.

\bibitem[Saharia et~al.(2022)Saharia, Chan, Saxena, Li, Whang, Denton,
  Ghasemipour, Ayan, Salimans, Ho, Fleet, and Norouzi]{saharia2022imagen}
Chitwan Saharia, William Chan, Saurabh Saxena, Lala Li, Jay Whang, Emily
  Denton, Seyed Kamyar~Seyed Ghasemipour, Burcu~Karagol Ayan, Tim Salimans,
  Jonathan Ho, David~J Fleet, and Mohammad Norouzi.
\newblock Imagen: Text-to-image diffusion models.
\newblock \emph{arXiv preprint arXiv:2205.11487}, 2022.

\bibitem[Schuhmann et~al.(2022)Schuhmann, Beaumont, Vencu, Gordon, Wightman,
  Cherti, Coombes, Katta, Mullis, Wortsman, Schramowski, Kundurthy, Crowson,
  Schmidt, Kaczmarczyk, and Jitsev]{laion5b}
Christoph Schuhmann, Romain Beaumont, Richard Vencu, Cade Gordon, Ross
  Wightman, Mehdi Cherti, Theo Coombes, Aarush Katta, Clayton Mullis, Mitchell
  Wortsman, Patrick Schramowski, Srivatsa Kundurthy, Katherine Crowson, Ludwig
  Schmidt, Robert Kaczmarczyk, and Jenia Jitsev.
\newblock {LAION-5B:} an open large-scale dataset for training next generation
  image-text models.
\newblock In \emph{NeurIPS}, 2022.

\bibitem[Shi et~al.(2023)Shi, Xiong, Lin, and Jung]{shi2023instantbooth}
Jing Shi, Wei Xiong, Zhe Lin, and Hyun~Joon Jung.
\newblock Instantbooth: Personalized text-to-image generation without test-time
  finetuning.
\newblock \emph{arXiv preprint arXiv:2304.03411}, 2023.

\bibitem[Sohl-Dickstein et~al.(2015)Sohl-Dickstein, Weiss, Maheswaranathan, and
  Ganguli]{sohl2015deep}
Jascha Sohl-Dickstein, Eric Weiss, Niru Maheswaranathan, and Surya Ganguli.
\newblock Deep unsupervised learning using nonequilibrium thermodynamics.
\newblock In \emph{International conference on machine learning}, pages
  2256--2265. PMLR, 2015.

\bibitem[Song et~al.(2020{\natexlab{a}})Song, Meng, and
  Ermon]{song2020denoising}
Jiaming Song, Chenlin Meng, and Stefano Ermon.
\newblock Denoising diffusion implicit models.
\newblock \emph{arXiv preprint arXiv:2010.02502}, 2020{\natexlab{a}}.

\bibitem[Song et~al.(2020{\natexlab{b}})Song, Sohl-Dickstein, Kingma, Kumar,
  Ermon, and Poole]{song2020score}
Yang Song, Jascha Sohl-Dickstein, Diederik~P Kingma, Abhishek Kumar, Stefano
  Ermon, and Ben Poole.
\newblock Score-based generative modeling through stochastic differential
  equations.
\newblock \emph{arXiv preprint arXiv:2011.13456}, 2020{\natexlab{b}}.

\bibitem[Wang et~al.(2004)Wang, Bovik, Sheikh, and Simoncelli]{ssim}
Zhou Wang, Alan~C Bovik, Hamid~R Sheikh, and Eero~P Simoncelli.
\newblock Image quality assessment: from error visibility to structural
  similarity.
\newblock \emph{IEEE transactions on image processing}, 13\penalty0
  (4):\penalty0 600--612, 2004.

\bibitem[Wei et~al.(2023)Wei, Zhang, Ji, Bai, Zhang, and Zuo]{wei2023elite}
Yuxiang Wei, Yabo Zhang, Zhilong Ji, Jinfeng Bai, Lei Zhang, and Wangmeng Zuo.
\newblock Elite: Encoding visual concepts into textual embeddings for
  customized text-to-image generation.
\newblock In \emph{Proceedings of the IEEE/CVF International Conference on
  Computer Vision}, pages 15943--15953, 2023.

\bibitem[Wu et~al.(2023)Wu, Huang, and Wei]{wu2023mole}
Xun Wu, Shaohan Huang, and Furu Wei.
\newblock Mole: Mixture of lora experts.
\newblock In \emph{The Twelfth International Conference on Learning
  Representations}, 2023.

\bibitem[Zhang et~al.(2023{\natexlab{a}})Zhang, Rao, and
  Agrawala]{zhang2023adding}
Lvmin Zhang, Anyi Rao, and Maneesh Agrawala.
\newblock Adding conditional control to text-to-image diffusion models.
\newblock In \emph{Proceedings of the IEEE/CVF International Conference on
  Computer Vision}, pages 3836--3847, 2023{\natexlab{a}}.

\bibitem[Zhang et~al.(2018)Zhang, Isola, Efros, Shechtman, and
  Wang]{zhang2018unreasonable}
Richard Zhang, Phillip Isola, Alexei~A Efros, Eli Shechtman, and Oliver Wang.
\newblock The unreasonable effectiveness of deep features as a perceptual
  metric.
\newblock In \emph{Proceedings of the IEEE conference on computer vision and
  pattern recognition}, pages 586--595, 2018.

\bibitem[Zhang et~al.(2023{\natexlab{b}})Zhang, Han, Ghosh, Metaxas, and
  Ren]{zhang2023sine}
Zhixing Zhang, Ligong Han, Arnab Ghosh, Dimitris~N Metaxas, and Jian Ren.
\newblock Sine: Single image editing with text-to-image diffusion models.
\newblock In \emph{Proceedings of the IEEE/CVF Conference on Computer Vision
  and Pattern Recognition}, pages 6027--6037, 2023{\natexlab{b}}.

\bibitem[Zhu et~al.(2021)Zhu, Feng, Shen, Zhao, Zha, Zhou, and
  Chen]{zhu2021low}
Jiapeng Zhu, Ruili Feng, Yujun Shen, Deli Zhao, Zheng-Jun Zha, Jingren Zhou,
  and Qifeng Chen.
\newblock Low-rank subspaces in gans.
\newblock \emph{Advances in Neural Information Processing Systems},
  34:\penalty0 16648--16658, 2021.

\end{thebibliography}
